\documentclass{article}

\usepackage{arxiv}

\usepackage[utf8]{inputenc} 
\usepackage[T1]{fontenc}    
\usepackage{hyperref}       
\usepackage{url}            
\usepackage{booktabs}       
\usepackage{amsfonts}       
\usepackage{nicefrac}       
\usepackage{microtype}      
\usepackage{lipsum}

\usepackage{amsmath, amssymb}
\usepackage{graphicx}
\usepackage{hyperref}
\usepackage{booktabs}
\usepackage{enumitem}
\usepackage{todonotes}
\usepackage{comment}
\usepackage{subcaption}
\usepackage{float}
\usepackage{xcolor}

\title{Multi-Objective Molecular Generation with Frequency-Controlled Evolutionary Dynamics}
\author{
Elia Colleoni \\
Clean Energy Research Center (CERP)\\
Physical Science and Engineering (PSE) Division \\
King Abdullah University of Science and Technology\\
Thuwal, Saudi Arabia 23955 \\
\texttt{elia.colleoni@kaust.edu.sa}
\And
Paolo Guida \\
Clean Energy Research Center (CERP)\\
Physical Science and Engineering (PSE) Division \\
King Abdullah University of Science and Technology\\
Thuwal, Saudi Arabia 23955 \\
\texttt{paolo.guida@kaust.edu.sa}
\And
 Didier Barradas-Bautista \\
  Kaust Visualization Lab (KVL)\\ 
  Core Lab Division\\
  King Abdullah University of Science and Technology\\
  Thuwal, Saudi Arabia 23955 \\
  \texttt{didier.bautista@kaust.edu.sa} \\
\And 
William L. Roberts \\
Clean Energy Research Center (CERP)\\
Physical Science and Engineering (PSE) Division \\
King Abdullah University of Science and Technology\\
Thuwal, Saudi Arabia 23955 \\
\texttt{william.roberts@kaust.edu.sa}
}

\date{\today}

\begin{document}

\maketitle

\begin{abstract}
Molecule generation methods that leverage generative models have been successfully applied to drug discovery. However, they often require extensive pre-training, suffer statistical biases in the training data, and might  suffer from limited interpretability of generated chemical structures. In this work, we introduce SpectralMol, an algorithm based on evolutionary computation that processes chemical structures as a compact matrix of Fourier coefficients, projected onto a fixed basis to generate position-wise latent vectors for SELFIES decoding. The NSGA-II algorithm enforces diversity and enable separate objective functions rather than collapsed objectives into a scalar reward. The quality of the algorithm was tested against standardized benchmarks. 
The results show comparable aggregate benchmark performance with a task-dependent profile: SpectralMol is strongest on several multi-parameter optimization tasks.
The same benchmark was used to perform an ablation study to demonstrate the advantages of a structured latent matrix. Finally, method was tested on a realistic ClpP-targeted drug-discovery benchmark, comparing it with the reinforcement-learning-based model under a fixed oracle-call budget. SpectralMol generates more docking hits and more diverse scaffolds while maintaining competitive physicochemical properties.
\textcolor{black}{The representation adopted in this work can cleanly separates scaffold-level modifications from localized substructure variations, as the former occur with perturbations of low-frequency Fourier modes and the latter with perturbations of high-frequency Fourier modes.} 
The results support the evidence that frequency-controlled evolutionary dynamics provide an interpretable, efficient, and training-free route to multi-objective molecular design.

\end{abstract}

\section{Introduction}
The process of discovering new molecules, while of tremendous importance, is often time-consuming and resource-intensive, as it generally involves synthesizing and testing several candidates. This is especially true in healthcare, where the estimated mean research and development cost for a new drug, considering only direct expenses, reached USD 368.6 million in 2019 \cite{mulcahy2025use}.
One way to mitigate risks is to test molecules similar to existing drugs \cite{eckert2007molecular,maggiora2014molecular}. This approach increases the chances of success but limits the opportunity to explore a wider chemical space. The advantages of virtual screening (VS) and, in general, of computer-aided drug discovery (CADD) have been conceptualized in the 1970s and have had a widespread impact since 1981 \cite{bleicher2003hit,van2007computer}. In recent years, methods that leverage artificial intelligence, particularly generative models, have revolutionized the field of materials discovery \cite{sanchez2018inverse,gomez2018automatic}.
While certainly powerful, generative algorithms (GAs) are generally dependent on the training dataset, thereby limiting the model's ability to explore unknown regions of chemical space. GAs can be broadly divided into four types: Variational autoencoder (VAE) based latent-space methods, reinforcement learning (RL) frameworks, transformer-based generative models, and evolutionary algorithms operating on molecular representations. 

In the first approach, VAEs learn continuous latent representations of discrete molecular structures. JT-VAE \cite{jin2018junction} represents molecules as junction trees of substructures, guaranteeing validity but requiring extensive pre-training and struggling beyond the training distribution.
LIMO \cite{eckmann2022limo}, pairs SELFIES with gradient-based latent optimization for 100\% molecular validity, but its reliance on gradients precludes non-differentiable objectives and risks local optima. More recently, LEOMol \cite{leomol2024} combined a ZINC250k-pretrained VAE with evolutionary computation on a 1024-dimensional latent space, resulting in the ability to handle non-differentiable oracles but remaining data-dependent. 
The second group leverages RL, framing the problem of generating molecules as a sequence of decisions that allows for direct optimization of non-differentiable objectives. MolDQN \cite{zhou2019optimization} uses double Q-learning on chemically valid actions, ensuring valid structures with no pre-training. Conversely, GCPN \cite{you2018graph} adopts policy gradients but requires pre-training and suffers from instability.
The third group, transformer-based generative algorithm, includes MolGPT \cite{bagal2021molgpt} and its various declinations cMolGPT \cite{wang2023cmolgpt}, MTMol-GPT \cite{tang2024mtmolgpt}, and T5-based T5MolGe \cite{xu2025screening}. These models treat molecule strings like text and are generated via next-token prediction, but, of course, they rely on extensive pre-training.

Aside from GAs, methods such as evolutionary algorithms (EAs) have played an important role in drug discovery. The latter leverages heuristic optimization, starting from a specific or randomly generated pool of molecules and gradually creating candidates that increasingly better fit a set of objective functions, identifying an optimal molecule. EAs overcome the limitations of GAs in terms of dataset dependence by providing greater molecular diversity and the flexibility to tailor objective functions to specific properties.
Despite their advantages and widespread adoption, EAs also have drawbacks; for example, during molecule generation, mutation and crossover mechanisms may yield unfeasible structures, an issue recently mitigated by the adoption of Self-Referencing Embedded Strings (SELFIES)\cite{Krenn_2020,nigam2021janus,nigam2022parallel,homberg2024optimized}.
In addition, some authors also proposed to convert molecules into a continuous latent-space representation as an effective way to smooth the search landscape and reduce the discontinuities introduced by direct graph or string-level edits \cite{sousa2021combining,grantham2022deep,liu2024multi}.
In most of previous works, the latent representation is either learned end-to-end by a variational autoencoder or treated as an unstructured real-valued vector on which evolutionary operators act directly, without imposing any regularity across sequence positions.

In this work, we take a different route and evolve molecules in a frequency-parameterized latent space. Rather than evolving the latent sequence itself, the genotype is a matrix of Fourier coefficients that is projected through a fixed basis to produce position-wise latent vectors used by the decoder. \textcolor{black}{This construction is designed so that low-frequency coefficients control coherent, large-scale changes across the sequence. In contrast, high-frequency coefficients act locally, yielding a structured form of exploration in which scaffold-level and substructure-level variations are disentangled by construction. This approach allows for combining the so-called "scaffold-hopping" \cite{bohm2004scaffold} with lead optimization \cite{temml2021structure}.}

Fourier decompositions of fitness functions and genotypes have long been studied in the theory of genetic algorithms \cite{kosters1999fourier}, and more recently, Fourier-parameterized latent sequences have been shown to provide smooth, structured representations of trajectories in deep generative models for robot motion learning \cite{li2024fld}. Our contribution is to bring this class of representation into molecular design, to combine it with a fixed, deterministic basis that removes the need for any pre-training stage, and to couple it with SELFIES-constrained decoding and a multi-objective evolutionary algorithm (NSGA-II in this case \cite{deb2000fast}). The resulting framework, referred to as \emph{SpectralMol} in the remainder of the paper, is therefore training-free, operates in a compressed and interpretable parameter space, and natively supports Pareto optimization over multiple chemical objectives, three features that, to the best of our knowledge, have not previously been combined in the molecular generation literature.

\section{Software description}
As previously mentioned, we apply an evolutionary algorithm to a matrix of coefficients named 
 $\Theta \in \mathbb{R}^{M \times D}$, where $M$ denotes the truncated number of Fourier modes and $D$ is the latent dimension of the decoder embedding space. All evolutionary operators, including both crossover and mutation, act exclusively on $\Theta$, manipulating the Fourier coefficients to generate molecular diversity.

A shared basis matrix $\Phi \in \mathbb{R}^{L \times M}$ provides the structural foundation for mapping sequence positions to Fourier features, where $L$ is the maximum decoded token length. In this work, $\Phi$ is constructed as a real-valued truncated trigonometric Fourier basis. Let $K$ denote the maximum retained harmonic. The number of Fourier modes is then
\[
M = 1 + 2K,
\]
corresponding to one constant mode, $K$ cosine modes, and $K$ sine modes. Using the zero-based sequence index $n=i-1$, with $i=1,\dots,L$, the entries of $\Phi$ are defined as
\[
\Phi_{i,0}=1,
\]
\[
\Phi_{i,k}
=
\cos\left(\frac{2\pi k n}{L}\right),
\qquad k=1,\dots,K,
\]
and
\[
\Phi_{i,K+k}
=
\sin\left(\frac{2\pi k n}{L}\right),
\qquad k=1,\dots,K.
\]
Thus, each row of $\Phi$ encodes the values of all retained real Fourier modes at the corresponding sequence position, while each column corresponds to a fixed Fourier basis function evaluated across the sequence. The basis remains constant throughout the optimization process and is not updated nor learned. 

Finally, the phenotypic representation emerges from the matrix product
\[
Z = \Phi\Theta \in \mathbb{R}^{L \times D}.
\]
Each row $Z_i$ serves as the latent vector that drives token selection at the corresponding position $i$ in the sequence. This construction ensures that the evolved Fourier coefficients in $\Theta$ are projected across all sequence positions via the fixed basis $\Phi$, yielding a continuous, structured latent trajectory that governs molecular generation. \textcolor{black}{Low-frequency coefficients induce coherent changes across many sequence positions, whereas higher-frequency coefficients introduce more localized variations in the decoded latent trajectory.}

\begin{figure}[h]
    \centering
    
    \includegraphics[width=\textwidth]{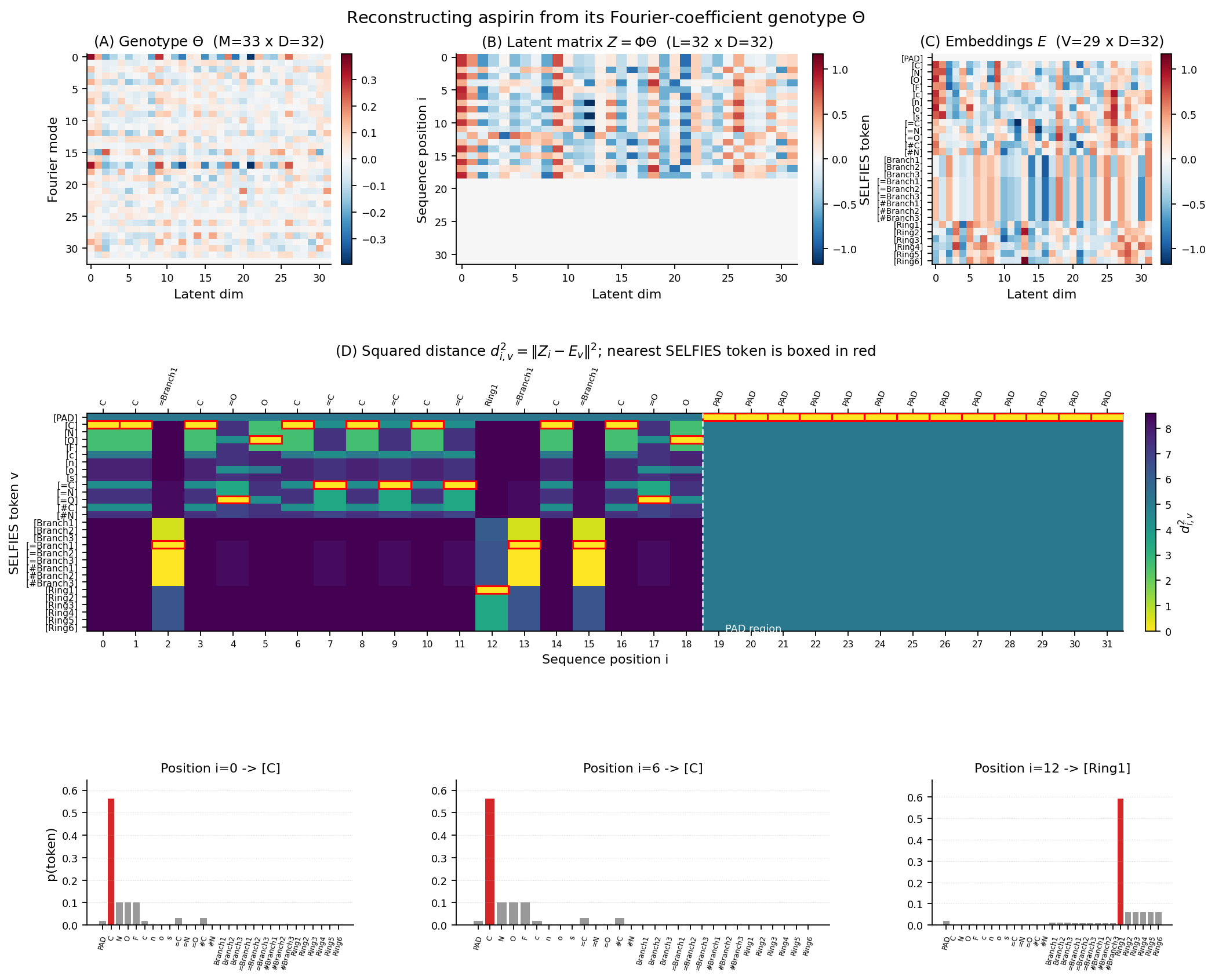}
    \caption{Schematic of the genotype-to-molecule reconstruction pipeline, illustrated for aspirin (acetyl salicylic acid).
    \textbf{(A)} The genotype $\Theta \in \mathbb{R}^{M \times D}$ ($M = 33$ Fourier modes, $D = 32$ latent dimensions) is the only object the evolutionary loop manipulates.
    \textbf{(B)} The phenotypic latent matrix $Z = \Phi\Theta \in \mathbb{R}^{L \times D}$ is obtained by projecting $\Theta$ through the fixed Fourier basis $\Phi$ ($L = 32$ sequence positions); horizontal banding reflects the smoothness imposed by the low-frequency basis, and the near-zero region from row 15 onward corresponds to the unused PAD positions.
    \textbf{(C)} The fixed token embedding table $E \in \mathbb{R}^{V \times D}$ ($V = 23$ tokens), constructed deterministically from chemical features via QR-orthonormal projection.
    \textbf{(D)} Squared Euclidean distances $d^{2}_{i,v} = \lVert Z_i - E_v \rVert^{2}$ between every position and every token. These distances are converted into token logits by negation and temperature scaling, $-d^{2}_{i,v}/T$, so that closer embeddings receive higher decoding probability. Constraint masks then set the probability of invalid tokens to zero before the final token is sampled from the remaining normalized distribution. The dashed white line marks the boundary between the molecule region and the PAD region, where valid termination positions collapse to \texttt{[PAD]}.}
    \label{fig:theta-to-molecule}
\end{figure}

Figure~\ref{fig:theta-to-molecule} illustrates the blocks that map a Fourier-coefficient genotype into a decoded SELFIES sequence, using acetyl salicylic acid (aspirin) as a reconstruction example.
Here, $L=32$ sequence positions and retain $K=16$ harmonics, yielding $M=1+2K=33$ Fourier coefficient rows. Therefore, $M$ is treated as a fixed truncation hyperparameter rather than being selected adaptively by an information criterion. We found that setting $L=32$ allows to retain enough expressivity while also retaining computational efficiency. 

\paragraph{Distance-based token probabilities}
The decoder operates on a token embedding table $E \in \mathbb{R}^{V \times D}$, where $V$ denotes the vocabulary size. For each sequence position $i$, the system computes the squared Euclidean distance between the position-specific latent vector and every token in the embedding matrix,
\[
d^2_{i,v} = \| Z_i - E_v \|^2 .
\]
These distances define token logits by negation and temperature scaling,
\[
\ell_{i,v} = -d^2_{i,v}/T,
\]
where $T>0$ controls the sharpness of the distribution. Closer token embeddings therefore receive larger logits and higher decoding probability. Before token selection, a constraint mask $m_i(v)\in\{0,1\}$ is applied to suppress structurally invalid tokens. 
Thus, any token that violates a decoding constraint has probability exactly zero, whereas valid tokens are sampled according to their distance-dependent probabilities.

\paragraph{Constraint-aware decoding}
SpectralMol adopts a masking procedure to enforce the validity of the generated structures. As the decoder associates a probability distribution with each sequence position, tokens that would violate the decoding constraints are assigned zero probability. 
The constraints are set as follow:
at decoding step $t$, we construct a binary mask $m_t\in\{0,1\}^{|\mathcal V|}$ from parser state $s_t=(\ell_t,b_t,\mathcal R_t,\nu_t)$, where $\ell_t$ is current length, $b_t$ is open-branch depth, $\mathcal R_t$ is the set of unmatched ring indices, and $\nu_t$ is remaining local valence. A token $k$ is suppressed by setting $m_t(k)=0$ if it violates any rule: \texttt{[PAD]} is masked unless $\ell_t\ge L_{\min}$, $b_t=0$, and $\mathcal R_t=\varnothing$; branch-open tokens are masked when $\nu_t=0$ or $b_t=B_{\max}$; branch-close tokens are masked when $b_t=0$; ring-open tokens are masked when no valence is available; ring-close tokens \texttt{[Ring$i$]} are masked unless ring $i$ is currently open and the implied bond is valence-compatible at both endpoints; and atom/bond tokens are masked whenever they exceed valence constraints. For macro tokens, masking is applied \emph{after macro expansion} by simulating each constituent token and suppressing the macro if any constituent step is invalid. If $\sum_k m_t(k)=0$, we deterministically unmask \texttt{[PAD]} to force termination and flag the sample for regeneration. This process results in generated sequences that are substantially less likely to contain malformed or chemically incompatible token patterns that would otherwise fail during the SELFIES-to-SMILES conversion (required for objective functions evaluation) or downstream molecular sanitation.

\paragraph{Vocabulary and embedding matrix}
The vocabulary is defined over SELFIES tokens and is extended with chemically meaningful macro tokens representing common rings, heterocycles, functional groups, and drug-like scaffold motifs.
The vocabulary is defined empirically and remains user-configurable: it can be extended, constrained, or tuned for different molecular design scenarios. This allows the representation to emphasize specific scaffolds and/or functional groups when targeting a particular chemical space, or alternatively to be expanded toward a more comprehensive coverage of chemically relevant tokens.
Token IDs are converted to their corresponding token strings through a lookup table. IDs corresponding to macro-tokens, are expanded into their constituent SELFIES tokens, and the resulting list is concatenated to form the final SELFIES string representation of the molecule.
We report the entire vocabulary in the supplementary material.

The embedding matrix \(E \in \mathbb{R}^{V \times D}\) is then constructed from structured token descriptors rather than from purely random embeddings. For each token ID, the implementation builds an interpretable feature vector that encodes its token class, such as padding, aliphatic atom, aromatic atom, bond, branch, ring, or macro token, together with element identity, chemical/structural flags, and ring-index information when applicable. These feature vectors are stacked and projected into the \(D\)-dimensional latent space using a seeded random orthogonal projection, followed by normalization and rescaling, with the padding embedding set to zero. Consequently, tokens are not placed arbitrarily in the latent space; chemically similar tokens are mapped to related regions because their embeddings are derived from shared structured features, while the random projection only determines the orientation of the final coordinate system. \textcolor{black}{This helps in modulating the research trajectory into scaffold-oriented modification, or more localized lead optimization.}

\paragraph{Crossover}

Recombination operations mix Fourier coefficients from genotypes to produce offspring. The matrix $\Theta$ is treated as a collection of row vectors corresponding to individual Fourier modes, and crossover operators, either uniform or segment-based, are applied to these rows to generate the offspring coefficient matrix $\Theta^{\mathrm{child}}$. The choice of crossover strategy determines how parental frequency components are blended: uniform crossover produces fine-grained mixtures, while segment crossover preserves larger coherent blocks of Fourier modes from each parent.

\paragraph{Mutation}

We employ a mutation strategy that acts on the matrix $\Theta$ and combines several complementary mechanisms to balance exploration and stability. First, we apply masked Gaussian noise to randomly selected coefficient subsets, introducing local perturbations while preserving the majority of the structure. Second, we resample Fourier modes occasionally, thereby enabling greater variability and expanding the exploration state. 
The third mechanism we have implemented is a progressive reduction in mutation intensity over generations, a feature that helps preserve suitable scaffolds and facilitates convergence. The entire mutation process is constrained by coefficient clipping and normalization.

The Fourier structure of the representation profoundly influences the phenotypic effects of mutations. \textcolor{black}{Perturbations to low-frequency modes induce coherent shifts in the latent vectors $Z_i$ across many sequence positions simultaneously, producing large-scale structural changes in the decoded molecule. Modifications to higher-frequency modes, in contrast, produce more localized variation, affecting smaller segments of the sequence and enabling fine-tuning of molecular substructures. This frequency-dependent behavior promotes diverse yet structured exploration of the molecular design space.}

\paragraph{Mapping molecular structures}

The complete transformation from genotype to evaluated molecule follows a distance-based probabilistic decoding pipeline. 
Each individual in the population undergoes the following sequence of operations:
\[
\Theta \xrightarrow{\ \Phi\ }\ Z\ (\zeta)\ \xrightarrow{\ D_{i,v} = \| Z_i - E_v \|^2\ }\ \text{token IDs}\ \xrightarrow{\ \text{lookup}\ }\ \text{SELFIES}\ .
\]
The Fourier coefficient matrix $\Theta$ is multiplied by the fixed basis $\Phi$ to yield the latent position matrix $Z$. 
This latent representation is fed into the decoder, which computes distance-based token probabilities against the embedding table $E$, applies validity masks to assign zero probability to invalid tokens, and samples a valid token ID at each sequence position.
The token IDs are mapped to their string representations and undergo macro-token expansion where applicable, resulting in a complete SELFIES string.

The computational complexity of this mapping is modest relative to the external oracle cost. For a single individual, projection from genotype to latent sequence scales as $O(LMD)$, nearest-neighbor token scoring scales as $O(LVD)$, and constraint masking plus token selection scales as $O(LV)$, where $L$ is sequence length, $M$ is the number of retained Fourier modes, $D$ is latent dimensionality, and $V$ is the decoder vocabulary size. SELFIES-to-SMILES conversion is approximately linear in sequence length, $O(L)$. For a population of size $P$, the non-oracle work per generation therefore scales as $O\!\left(P(LMD + LVD)\right)$, with the dominant wall-clock cost in docking-based experiments arising from oracle evaluation rather than from the latent representation itself.

From an implementation perspective, the pipeline can be summarized in four stages. First, the Fourier coefficient matrix is projected through the fixed basis to obtain the latent sequence matrix $Z$. Second, each position-specific latent vector is compared against every token embedding to obtain logits. 
Third, structural masks suppress invalid tokens, and a valid token ID is selected. Fourth, token IDs are converted to SELFIES with macro-token expansion where needed. 

\begin{figure}[h]
    \centering
    \includegraphics[width=\textwidth]{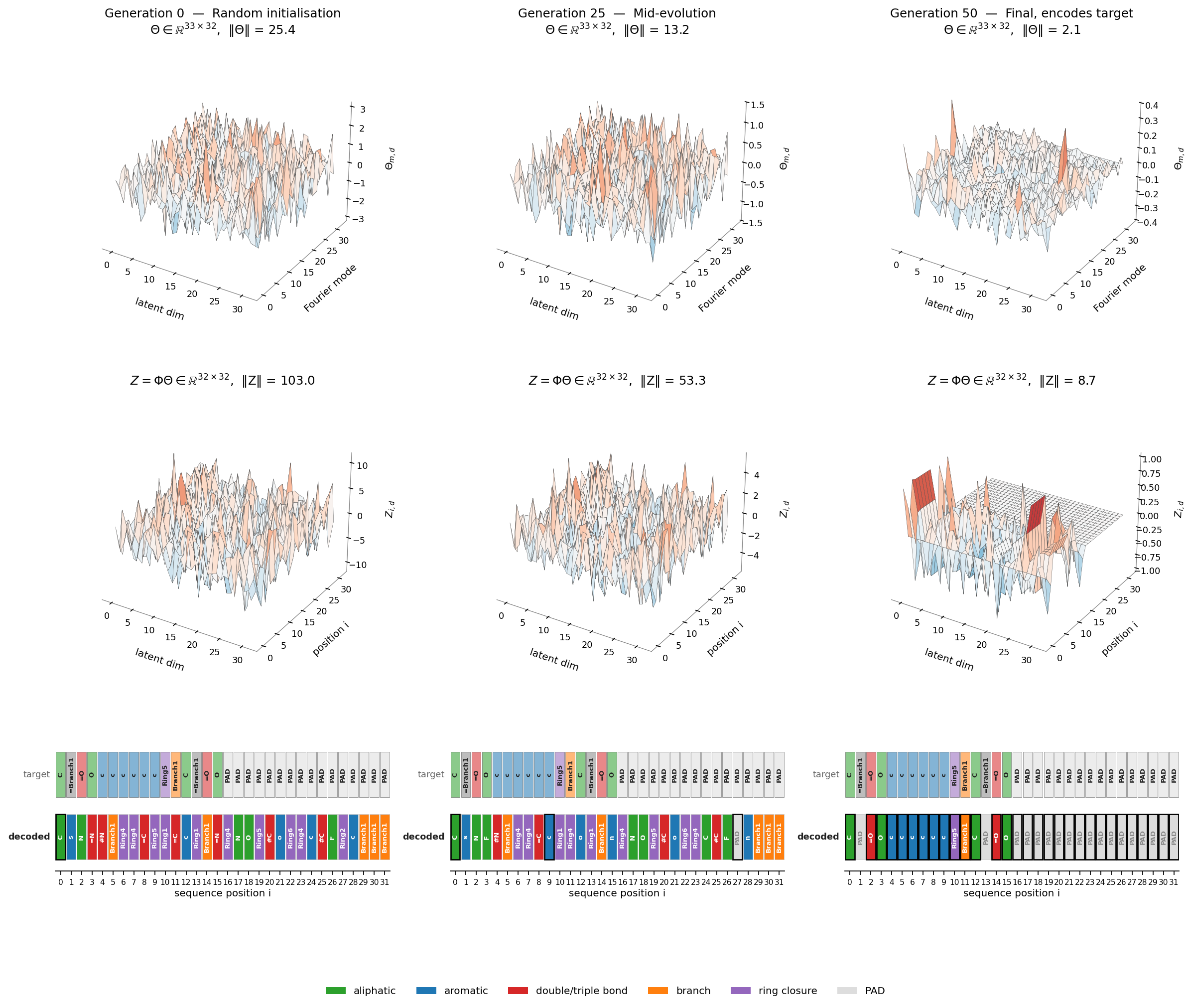}
    \caption{Co-evolution of the genotype $\Theta$, its phenotype $Z = \Phi\Theta$, and the decoded token sequence across three representative generations of the SpectralMol GA, illustrated for a target encoding aspirin (acetyl salicylic acid).
    \textbf{Top row:} the Fourier coefficient matrix $\Theta \in \mathbb{R}^{33 \times 32}$ at generation 0 (random initialization), generation 25 (mid-evolution), and generation 50 (final, encoding the target). Frobenius norm decreases from $\lVert \Theta \rVert = 25.4$ to $13.2$ to $2.1$ as the genotype converges onto the sparse low-frequency combination needed to reproduce the target embedding rows.
    \textbf{Middle row:} the corresponding latent matrix $Z = \Phi\Theta \in \mathbb{R}^{32 \times 32}$ (sequence position $\times$ latent dimension), with $\lVert Z \rVert$ shrinking from $103.0$ to $53.3$ to $8.7$ as the phenotype collapses from broadband noise toward the unit-norm subspace spanned by the embedding table $E$.
    \textbf{Bottom row:} the decoded SELFIES token sequence at each generation (\emph{decoded}) compared to the target (\emph{target}). Cells are coloured by token category (aliphatic, aromatic, double/triple bond, branch, ring closure, PAD); black borders mark positions where the decoded token matches the target. Match counts increase from $1/32$ at generation~0 to $3/32$ at generation~25 and $30/32$ at generation~50, illustrating how the frequency-controlled latent representation converges to a chemically meaningful sequence as evolution progresses.}
    \label{fig:theta-evolution}
\end{figure}

\paragraph{Optimization routine}
We employ NSGA-II for multi-objective optimization rather than scalarized reward functions or weighted sums.
NSGA-II maintains a population representing the Pareto frontier through non-dominated sorting and crowding distance maximization, naturally handling competing objectives (e.g., maximizing Quantitative Estimate of Drug-likeness, QED, while maintaining scaffold similarity) without requiring manual weight specification. This contrasts sharply with scalarized multi-objective RL and transformer-based conditional generation methods, both of which require task-specific engineering for each objective combination. The evolution process is represented in  Figure \ref{fig:theta-evolution}.
The figure provides a mechanistic illustration of how the evolutionary process modifies the Fourier-based genotype and how these changes propagate to the decoded molecular sequence. Across the three representative generations, the coefficient matrix \(\Theta\) evolves from a random, high-amplitude initialization toward a more structured and lower-norm representation. This convergence is reflected in the corresponding phenotype \(Z = \Phi\Theta\), whose initially noisy latent trajectory progressively collapses toward the region of the embedding space associated with valid SELFIES tokens. Importantly, the figure shows that optimization does not act directly on discrete molecular strings, but instead reshapes a continuous frequency-parameterized representation, which is then decoded position by position through the embedding table. The increasing agreement between the decoded sequence and the target sequence, from only a few matching tokens in the early generations to near-complete reconstruction in the final generation, illustrates how selection, crossover, and mutation jointly refine the spectra toward matching the target molecule.

\paragraph{Comparative landscape}

Table~\ref{tab:comparison} provides a systematic comparison of our approach against representative methods from each category. The comparison highlights fundamental architectural choices regarding training requirements, parameterization strategy, optimization framework, and computational characteristics.

\begin{table*}[t]
\centering
\caption{Comparison of molecular generation approaches across key dimensions.}
\label{tab:comparison}
\small
\begin{tabular}{@{}lp{1.7cm}p{2cm}p{1.8cm}p{1.5cm}p{1.7cm}p{2cm}@{}}
\toprule
\textbf{Method} & \textbf{Representation} & \textbf{Training Required} & \textbf{Optimization} & \textbf{Latent Dim.} & \textbf{Chemistry Priors} & \textbf{Key Limitation} \\
\midrule
JT-VAE \cite{jin2018junction} & Junction tree & VAE pre-training & Latent sampling & 56 & Implicit & Limited to training distribution \\
LIMO \cite{eckmann2022limo} & SELFIES & VAE pre-training & Gradient descent & 1024 & Implicit & Cannot use non-differentiable oracles \\
LEOMol \cite{leomol2024} & SELFIES & VAE pre-training (18 epochs) & GA/DE on latent vectors & 1024 & Implicit & Unstructured latent space, data-dependent \\
MolDQN \cite{zhou2019optimization} & SMILES graph & None (policy learning) & Q-learning & N/A & Explicit & Sequential construction, expensive RL \\
GCPN \cite{you2018graph}& Molecular graph & Discriminator + policy pre-training & Policy gradient + adversarial & N/A & Implicit & Pre-training required, adversarial instability \\
MolGPT \cite{bagal2021molgpt} & SMILES & Transformer pre-training & Conditional generation & Hidden (256-768) & Implicit & Data-driven, limited property control \\
cMolGPT \cite{wang2023cmolgpt} & SMILES & Transformer pre-training + fine-tuning & Conditional generation & Hidden (256-768) & Implicit & Requires target-specific fine-tuning \\
GB-GA \cite{jensen2019graph} & Molecular graph & None & GA on graphs & N/A & Explicit & Discontinuous jumps, no learned patterns \\
STONED \cite{nigam2021stoned}& SELFIES & None & String mutations & N/A & Explicit & Random walk, no smooth optimization \\
\midrule
\textbf{This work} & \textbf{SELFIES} & \textbf{None} & \textbf{NSGA-II on Fourier coefficients} & \textbf{$M\times D$ (1056)} & \textbf{Explicit} & \textbf{Hand-coded heuristics, no learned priors} \\

\bottomrule
\end{tabular}
\end{table*}

The approach presented in this work synthesizes insights from multiple methodological traditions while addressing key limitations of existing frameworks. 
(1) no neural network training is required (neither VAE pre-training nor policy network optimization), the method can be deployed immediately on new chemical spaces or property combinations. The deterministic embedding construction and Fourier basis calculation are one-time operations. During evolution, the primary computational costs are Fourier projection (matrix multiplication $\Phi\Theta$), distance calculations to the embedding table, and SELFIES decoding—all of which are orders of magnitude faster than training neural networks or performing sequential RL rollouts. This efficiency, combined with interpretability and multi-objective capability, positions the work as particularly suitable for early-stage drug discovery scenarios where exploration of diverse chemical scaffolds across multiple property criteria is paramount, and interpretability of design choices is valued alongside performance.
(2) in contrast to direct evolutionary algorithms operating on graphs or strings, we evolve in a smooth, structured latent space induced by Fourier parameterization. 
The genotype is not the latent sequence itself, but rather a matrix of Fourier coefficients which is evolved during optimization. 
This Fourier representation has several advantages. First, it dramatically reduces the parameter space compared to direct latent vector evolution. An example of this ca be seen in the comparison with LEOMol, where optimization acts on 80 $\times$ 1024 = 81,920 latent dimensions versus the only 33 $\times$ 32 = 1,056 Fourier coefficients of SpectrlMol.
\textcolor{black}{(3), it enforces smoothness: perturbations to low-frequency Fourier modes induce coherent changes across many sequence positions, enabling large-scale scaffold transformations, while high-frequency mode mutations produce localized refinements. This frequency-dependent behavior is mathematically guaranteed by the Fourier basis and provides structured exploration unavailable in unparameterized latent spaces.}
(4) our decoder architecture distinguishes itself from both learned decoders (VAEs, transformers) and simple nearest-neighbor schemes through the integration of explicit chemical heuristics constraints. 
Such constraints are not learned from data but rather encoded based on chemical knowledge and on common failure modes of stochastic molecular generators, specifically, the tendency to produce unclosed long chains or truncated structures. This hybrid approach combines the flexibility of probabilistic sampling with the reliability of rule-based constraints.


Table~\ref{tab:contributions} delineates the specific technical innovations of our work relative to the most closely related approaches: LEOMol (evolutionary optimization in VAE latent space), GCPN (RL-based graph generation), and MolGPT (transformer-based generation). The combination of these features—particularly zero-shot operation, Fourier smoothness guarantees, and native multi-objective optimization, represents a unique position in the design space of molecular generative models. While individual features appear in isolation across existing methods (e.g., LEOMol also uses evolutionary algorithms, GCPN also incorporates domain knowledge), their integration in our framework addresses a previously unoccupied niche: interpretable, training-free, multi-objective molecular optimization with mathematically structured exploration dynamics.

\begin{table*}[h]
\centering
\caption{Specific technical contributions distinguishing this work from closest methodological relatives.}
\label{tab:contributions}
\small
\begin{tabular}{@{}lcccc@{}}
\toprule
Feature & SpectralMol & LEOMol & GCPN & MolGPT \\
\midrule
Zero-shot (no pre-training) & \checkmark & $\times$ & $\times$ & $\times$ \\
Deterministic interpretable embeddings & \checkmark & $\times$ & $\times$ & $\times$ \\
Fourier-parameterized smooth latent space & \checkmark & $\times$ & N/A & N/A \\
Frequency-dependent exploration dynamics & \checkmark & $\times$ & $\times$ & $\times$ \\
Explicit chemical heuristics in decoder & \checkmark & $\times$ & Implicit & $\times$ \\
True multi-objective Pareto optimization & \checkmark & $\times$ & $\times$ & $\times$ \\
Compressed parameter space (1K vs 80K) & \checkmark & $\times$ & N/A & N/A \\
Non-differentiable oracle compatibility & \checkmark & \checkmark & \checkmark & $\times$ \\
Molecule-level generation (non-sequential) & \checkmark & \checkmark & $\times$ & $\times$ \\
\bottomrule
\end{tabular}
\end{table*}

\section{Results and discussion}

\paragraph{Benchmark} To evaluate the performance of the proposed algorithm, we adopted a two-stage benchmarking strategy designed to test general optimization performance and practical chemical relevance. In the first test case, we evaluated SpectralMol using a set of commonly used benchmarks, namely GuacaMol \cite{brown2019guacamol}, 
and MolExp(L) \cite{thomas2025test}, both implemented within the MolScore framework \cite{thomas2024molscore}. Each benchmark consists of multiple tasks, with each task corresponding to a distinct objective function that is independently optimized.
The score of each task is normalised to the interval $[0,1]$, and the benchmark overall result is given by the sum of individual tasks. The objective of the first test was to evaluate the performance against several different goal-directed tasks.

For the second test case, we opted for a more realistic drug discovery scenario, where multiple chemically relevant properties are optimized simultaneously. 
We compared against the published results of Saturn, the reinforcement-learning-based autoregressive molecular generative model recently introduced by Guo et al. \cite{guo2024saturn,guo2025directly}.
Overall, the two test cases are complementary: the first emphasizes standardized benchmark comparability, whereas the second probes the practical relevance of multi-objective optimization under more stringent design conditions.

\subsection{Test case 1: MolScore}

The rationale for this test case is that the selected benchmarks probe complementary aspects of molecular generation under both goal-directed and exploration-based settings. The GuacaMol \cite{brown2019guacamol} benchmarks provide standardized optimization tasks that are widely used to assess whether a generative model can efficiently identify molecules matching predefined structural or property-based objectives. In particular, the GuacaMol task set comprises 20 tasks, spanning rediscovery, similarity, scaffold hopping, median molecule generation, isomer-related objectives, and multi-parameter optimization. The MolExp(L) \cite{thomas2025test} benchmark family addresses a different aspect of molecular generation, the chemical exploration across multiple rewarding regions of chemical space. In the current evaluation setup, the MolExp benchmark comprises 5 tasks, whereas MolExpL represents a more demanding exploration regime comprising 4 tasks. In addition, two baseline-derived benchmark sets were considered: MolExp\_baseline and MolExpL\_baseline, each consisting of 12 task-specific variants over base-case objectives. Table~\ref{tab:molscore_testcase1} reports a summary of the benchmark families, task counts, and evaluation objectives considered in this test case.

\begin{table}[h!]
\centering
\caption{Summary of the benchmark families considered in Test case 1, the number of tasks, and the main type of task evaluated.}
\label{tab:molscore_testcase1}
\resizebox{0.8\textwidth}{!}{%
\begin{tabular}{p{3.0cm} c p{4.4cm} p{4.8cm}}
\hline
\textbf{Benchmark} & \textbf{\# Tasks} & \textbf{Main task type} & \textbf{What is evaluated} \\
\hline
GuacaMol & 20 & Goal-directed molecular optimization & Optimization toward predefined objectives, including rediscovery, similarity matching, scaffold hopping, median molecule generation, isomer design, and multi-parameter optimization. \\
MolExp & 5 & Multi-task chemical exploration & Discovery of high-reward molecules across AP bioactivity, AP, A2A, EGFR, and BACE1 objective settings. \\
MolExp\_baseline & 12 & Baseline exploration setting & Reference benchmark including task-specific variants over AP, A2A, BACE1, and EGFR objectives. \\
MolExpL & 4 & Large-scale chemical exploration & More demanding exploration setting including BACE1, A2A, A2A bioactivity, and AP objectives. \\
MolExpL\_baseline & 12 & Baseline large-scale exploration setting & Reference large-scale benchmark including task-specific variants over AP, A2A, BACE1, and EGFR objectives. \\
\hline
\end{tabular}%
}
\end{table}

The performance of SpectralMol was assessed by comparing the benchmark results with a well-known and established generative algorithm GraphGA \cite{jensen2019graph} (again part of the MolScore framework), using the version implemented in the repository \texttt{MolScore\_examples} \cite{molscore_examples}.
To ensure a fair head-to-head evaluation, both SpectralMol and GraphGA were launched through the same benchmarking harness with matched protocol settings.
Table~\ref{tab:budget_selected} summarizes the evaluation-budget definition and oracle accounting used in the selected-family comparison, including the fixed iteration limit, population and offspring settings, and the corresponding number of scored molecules per task for SpectralMol and GraphGA.
In the next paragraph, a task-by-task comparison across all benchmarks is presented, evaluating the differences between the SpectralMol model presented here and the GraphGA generative model.

In the next section, a detailed discussion for GuacaMol \cite{brown2019guacamol} benchmark is reported, results related to MolExp(L) \cite{thomas2025test} are reported in the supplementary material.

\begin{table}[h]
\centering
\caption{Budget/accounting clarification for the direct comparison runs.}
\label{tab:budget_selected}
\begin{tabular}{ll}
\hline
Setting & Value \\
\hline
Generation limit (reported) & 500 evolutionary iterations \\
Population size & 256 \\
Per-generation offspring/batch size & 451 \\
SpectralMol scored molecules per task & 225,756 \\
GraphGA scored molecules per task & 189,567--255,061 \\
\hline
\end{tabular}
\end{table}

\paragraph{Guacamol Benchmarks}

\begin{figure}[h]
    \centering
    \includegraphics[width=\linewidth]{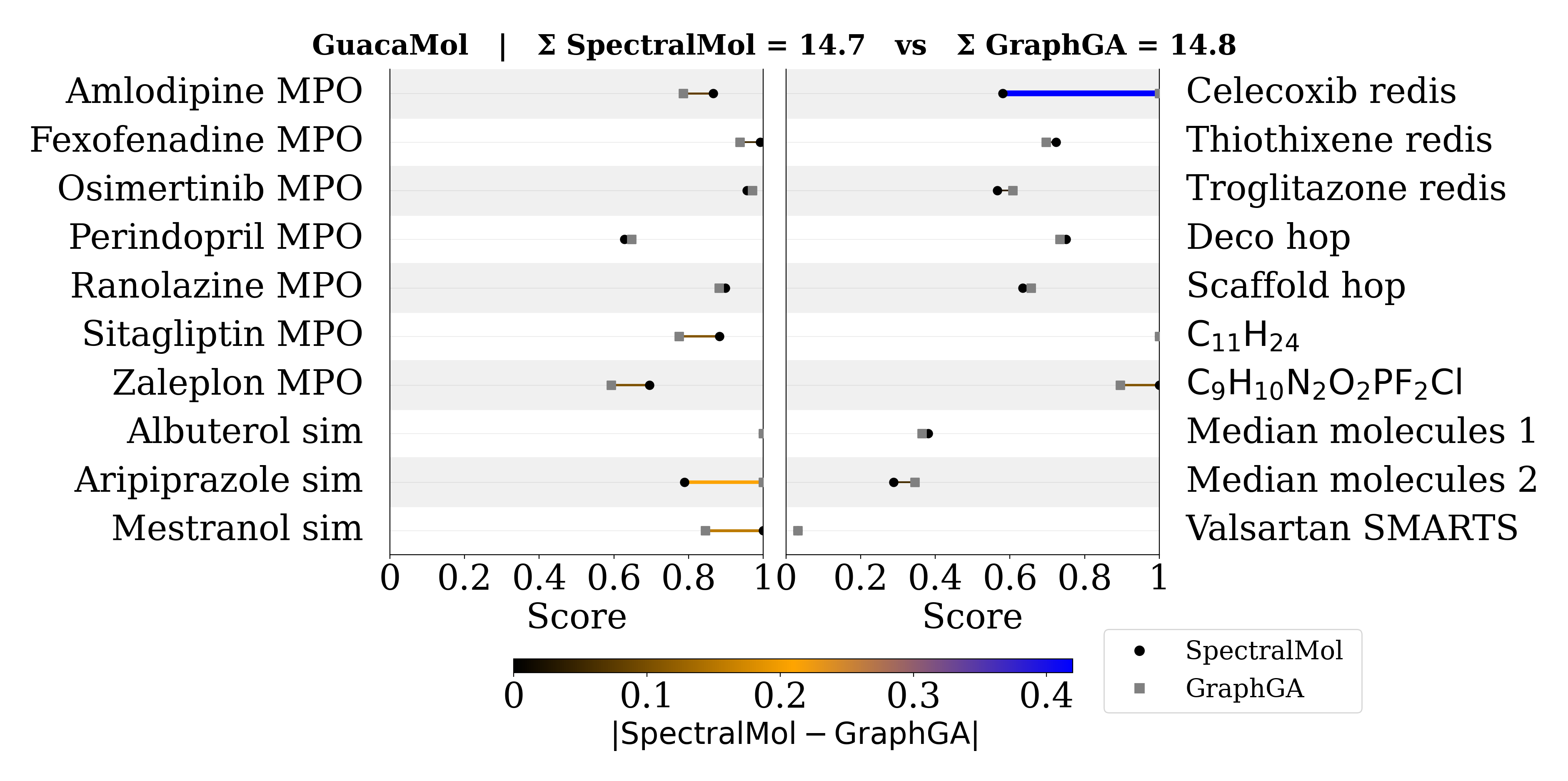}
    \caption{Task-by-task comparison between SpectralMol (blue) and GraphGA (grey) on the 20 tasks of the GuacaMol benchmark, grouped by category (multi-parameter optimization, similarity, rediscovery, isomer-related, hopping, median molecules, and SMARTS-constrained tasks). For each task, the horizontal segment connects the two methods, with the bar colour encoding the absolute score difference $|\mathrm{SpectralMol}-\mathrm{GraphGA}|$ according to the colourbar. The aggregated benchmark score ($\Sigma$, sum of per-task best scores) is reported in the panel header (14.7 for SpectralMol vs.\ 14.8 for GraphGA), indicating overall comparable optimization capacity with task-dependent strengths. Both methods used identical protocol settings (500 generations, population size 256, ChEMBL24-derived seed pool of size 2000, shared initial population).}
    \label{GucaMol_tasks}
\end{figure}

Figure~\ref{GucaMol_tasks} shows that the overall performance of the two models is comparable, with aggregate scoring (defined as the sum of the best score for each task) equal to 14.7 and 14.8 for SpectralMol and GraphGA, respectively. 
However, the difference between SpectralMol and GraphGA is strongly task-dependent. 
In the figure, tasks are grouped by family, starting with multi-parameter optimization (MPO) tasks. Here, SpectralMol is generally superior, with visible gains on Amlodipine\_MPO, Fexofenadine\_MPO, Sitagliptin\_MPO, and Zaleplon\_MPO, and a smaller advantage on Ranolazine\_MPO; by contrast, Osimertinib\_MPO and Perindopril\_MPO slightly favor GraphGA. 
The similarity (sim) and rediscovery (redis) objectives are reported. 
Regarding those tasks, the two methods show a mixed behavior: SpectralMol performs better on Mestranol\_similarity, 
GraphGA has a clear advantage on Aripiprazole\_similarity, and Albuterol\_similarity is essentially saturated for both methods.
On the other hand, GraphGA is clearly better on Celecoxib\_rediscovery, SpectralMol is slightly better on Thiothixene\_rediscovery, and Troglitazone\_rediscovery remains marginally in favor of GraphGA.
Among the remaining tasks, C11H24 is effectively solved equally well by both methods, while SpectralMol shows a modest advantage on C9H10N2O2PF2Cl. In the hopping objectives, SpectralMol is slightly better on Deco\_hop, whereas Scaffold\_hop shows a small advantage for GraphGA. For the median-molecule tasks, SpectralMol performs better on Median\_molecules\_1, while Median\_molecules\_2 slightly favors GraphGA. Finally, Valsartan\_smarts is challenging for both methods, with very low scores overall and only a minor difference between them. 

Consistent with these task-level trends, the total GuacaMol score remains nearly identical for the two approaches, indicating that the genotype--phenotype formulation preserves the overall optimization capability of the baseline while changing its task-specific behavior. Particularly, 
SpectralMol performs better on the MPO objectives, surpassing GraphGA in five of the seven MPO tasks.


Because GuacaMol includes twenty different optimization tasks, it is useful to complement the final benchmark score with two additional analyses: \mbox{(i)} the evolutionary trajectories of the best score during optimization and \mbox{(ii)} the score distributions of the top 200 molecules produced by SpectralMol and GraphGA. Together, these views clarify not only which method attains the highest final score, but also how optimization progresses and whether the advantage is limited to a few outliers or extends to the whole elite population. Additional task-level analyses for the MolExp, MolExpL, and Saturn benchmarks are provided in the Supplementary Material.

Figure~\ref{GucaMol_evolution} shows the best-score trajectory over 500 generations.
The trajectory analysis confirms that the two methods have very similar overall optimization capability, but markedly different task-dependent behaviors. Most tasks show rapid improvement in the first generations, followed by a plateau, indicating that both algorithms identify good regions of chemical space early and then perform limited refinement. 

\begin{figure}[h]
    \centering
    \includegraphics[width=\linewidth]{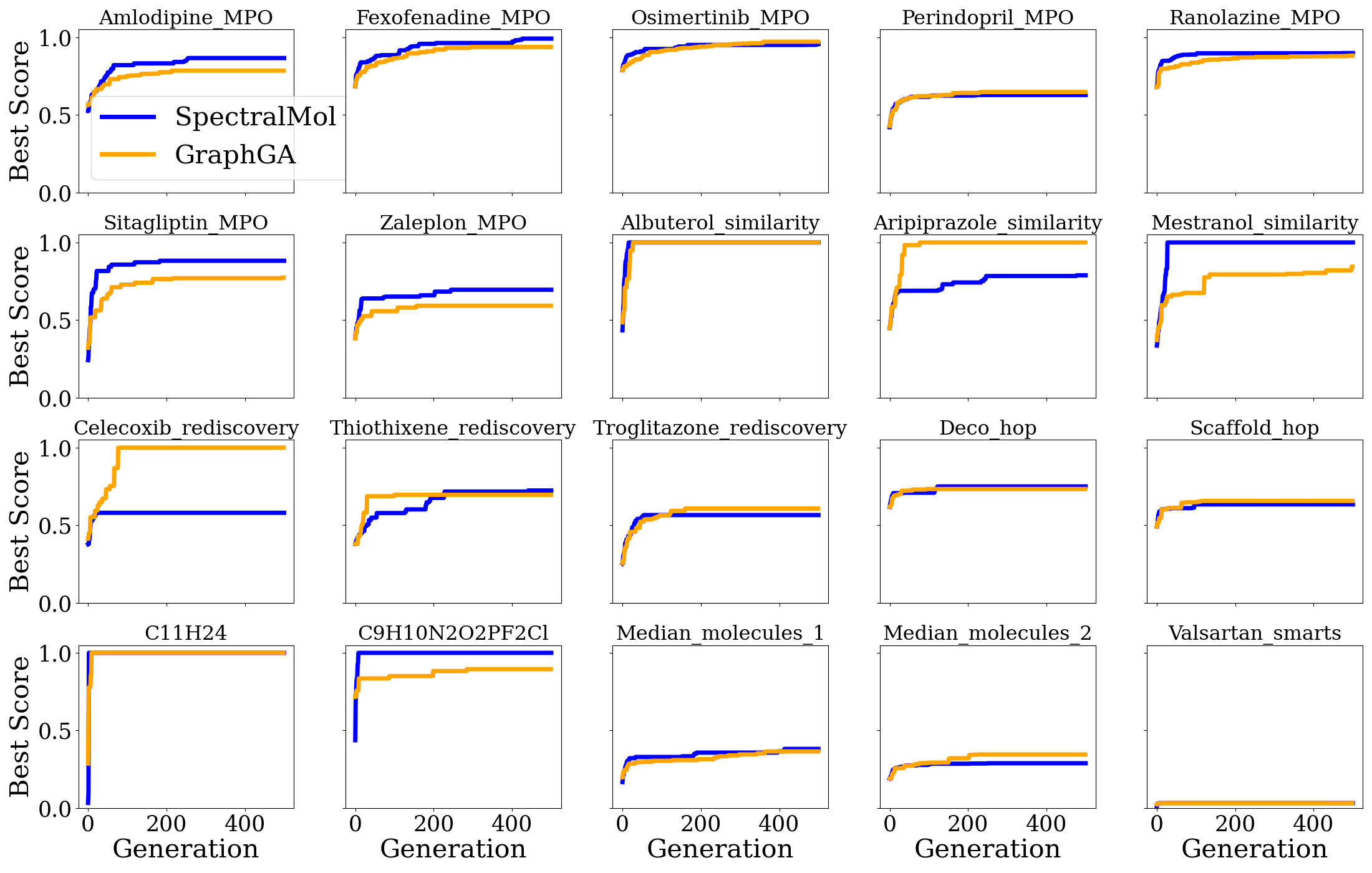}
    \caption{Best-score evolutionary trajectories over 500 generations for SpectralMol (blue) and GraphGA (orange) across the 20 GuacaMol tasks. Each panel reports the running best score within the population versus generations number for one task. Most tasks display rapid initial improvement followed by a plateau, indicating that both algorithms identify favorable regions of chemical space early in the run and devote later generations to refinement. Task-dependent differences are visible: SpectralMol shows clearer trajectories on multi-parameter optimization tasks (e.g., Amlodipine\_MPO, Sitagliptin\_MPO, Mestranol\_similarity), whereas GraphGA progresses faster on tasks requiring exact structural reconstruction (e.g., Aripiprazole\_similarity, Celecoxib\_rediscovery). 
    Both methods used identical protocol settings.}

    \label{GucaMol_evolution}
\end{figure}

\begin{figure}[h]
    \centering
    \includegraphics[width=\linewidth]{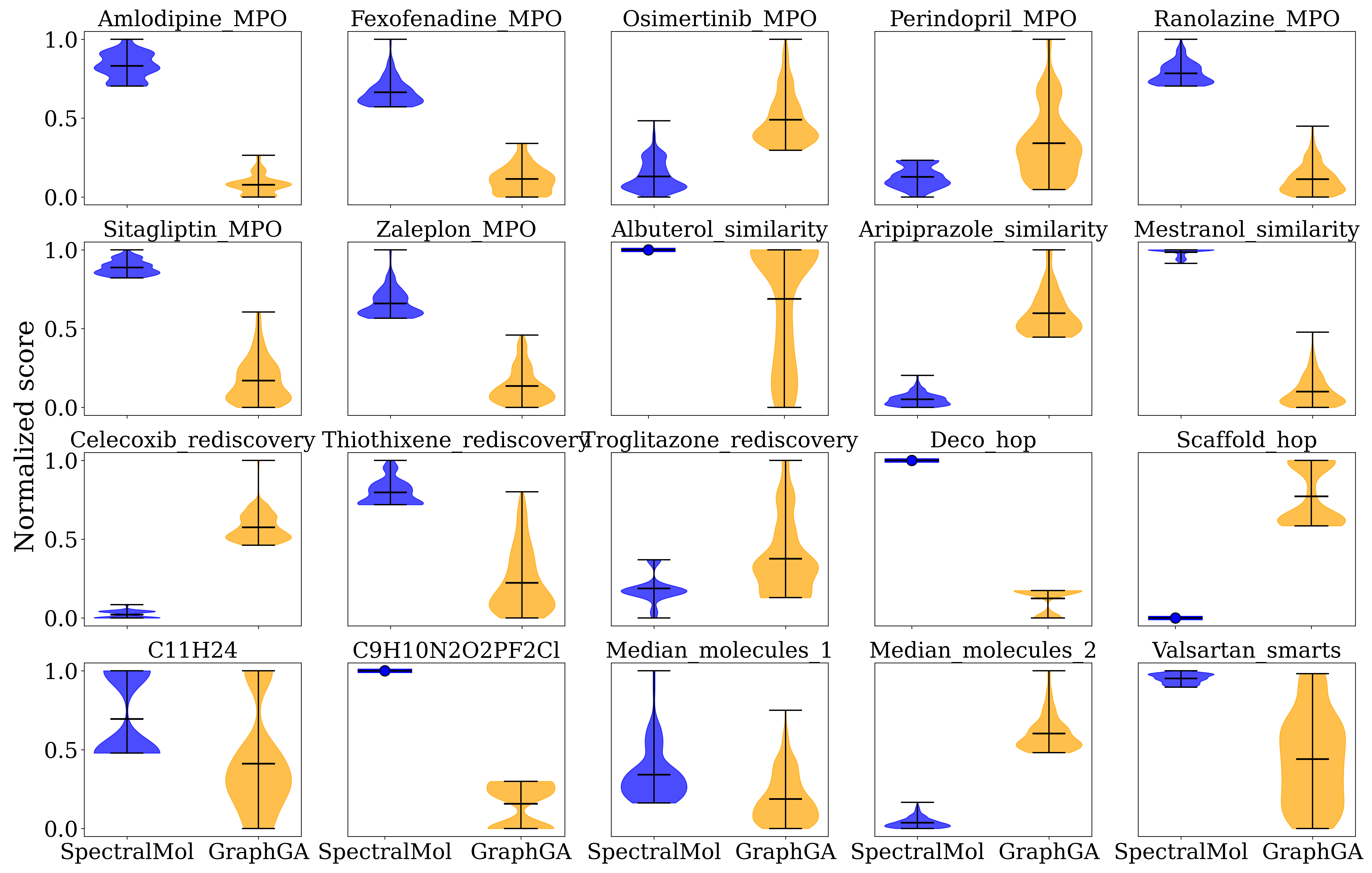}
    \caption{Score distributions of the top-200 molecules produced by SpectralMol (blue) and GraphGA (orange) on each of the 20 GuacaMol tasks. Violin plots show the kernel-density estimate of the normalized score across the elite population, with internal lines indicating the median and interquartile range. This population-level view complements the best-score trajectories of Figure~\ref{GucaMol_evolution}, clarifying whether a method's advantage on a given task originates from a few exceptional outliers or extends to the entire elite set. SpectralMol elite distributions are markedly right-shifted on the multi-parameter optimization tasks (Amlodipine\_MPO, Fexofenadine\_MPO, Sitagliptin\_MPO, Zaleplon\_MPO) and on Mestranol\_similarity. In several tasks (Albuterol\_similarity, Deco\_hop, Scaffold\_hop, C9H10N2O2PF2Cl) the SpectralMol violin collapses into a thin line, indicating highly concentrated elite populations.}
    \label{GucaMol_distribution}
\end{figure}

Figure~\ref{GucaMol_distribution} reports the score distributions of the top 200 molecules to provide a population-level view. The most informative cases are those in which the two methods differ not only in final score, but also in evolutionary trajectory and in the distribution of high-scoring solutions. 
Amlodipine\_MPO, Fexofenadine\_MPO, Sitagliptin\_MPO, and Zaleplon\_MPO consistently favor SpectralMol in both analyses: SpectralMol attains higher scores during evolution and its top-200 molecules are shifted toward better values (closer to 1), indicating that, on multi-property objectives, the genotype--phenotype formulation improves the overall quality of the search rather than merely producing a few exceptional outliers. An even more pronounced trend is observed for Mestranol\_similarity, where SpectralMol rapidly approaches near-optimal scores and concentrates most of its top solutions close to the optimum.
Conversely, Aripiprazole\_similarity and Celecoxib\_rediscovery are the clearest examples in favor of GraphGA. Here, GraphGA shows both superior trajectories and a population-wide right shift in the score distribution, suggesting that direct graph-level edits remain more effective when the objective requires reconstructing precise structural motifs or following a narrow path toward a target compound. Finally, Valsartan\_smarts is informative because both methods remain at very low scores and show little improvement over time. This is an indication that, under highly restrictive SMARTS-based constraints, the main bottleneck is likely the search operators themselves rather than the molecular representation.
Interesting to note is that, in a few tasks, the violin plots (Figure~\ref{GucaMol_distribution}) collapse into a line. This is most evident for Albuterol\_similarity Deco\_hop and Scaffold\_hop, and C$_9$H$_{10}$N$_2$O$_2$PF$_2$Cl, where SpectralMol produces a highly concentrated elite population with all molecules scoring the same value.

Overall, these results indicate that SpectralMol is particularly effective on constrained multi-objective optimization tasks, whereas GraphGA can still be preferable on problems that benefit from more direct structural reconstruction. 

\paragraph{GuacaMol frequency-mode ablation}

To isolate the role of frequency-mode choices in the active SpectralMol runner, we executed a GuacaMol ablation over four conditions (full-spectrum, high-only, low-only, random-matrix) using matched settings (20 tasks, 6 seeds, budget 20000 per task).
Table~\ref{tab:guacamol_ablation_overall_main} summarizes the condition-level outcomes from the finalized run reports.

\begin{table}[h!]
\centering
\caption{GuacaMol ablation summary across frequency-mode conditions. Values are aggregated over 20 tasks and 6 seeds per condition.}

\label{tab:guacamol_ablation_overall_main}
\small
\begin{tabular}{lcccccc}
\hline
Condition & Mean best score & $\Delta$ vs full & Mean elapsed (s) & Speedup vs full & Evals/s & $n_{\mathrm{failed}}$ \\
\hline
high-only & 0.7542 & +0.0050 & 1331.3 & 1.156$\times$ & 15.023 & 0 \\
full-spectrum & 0.7492 & 0.0000 & 1538.6 & 1.000$\times$ & 12.999 & 0 \\
low-only & 0.7367 & -0.0125 & 1548.1 & 0.994$\times$ & 12.919 & 0 \\
random-matrix & 0.6469 & -0.1023 & 1256.1 & 1.225$\times$ & 15.923 & 0 \\
\hline
\end{tabular}
\end{table}

The ablation confirms that \emph{high-only} slightly exceeds full-spectrum in mean best score (+0.0050) while also reducing elapsed time (1.156$\times$ speedup). \emph{Low-only} underperforms full-spectrum by -0.0125, and \emph{random-matrix} is fastest but shows the largest quality drop (-0.1023), highlighting a clear quality-throughput trade-off. 

These results confirm that high-frequency structure contributes most to quality-speed balance across all ablation and matrix configurations.

\paragraph{Computational Acceleration}

SpectralMol and GraphGA matrix runs were benchmarked against GuacaMol tasks across four configurations: baseline CPU, Dask CPU, RAPIDS V100, and Dask+RAPIDS V100. The latest results (Table~\ref{tab:matrix_acceleration}) show that Dask+RAPIDS on V100 achieves a 14.6\% reduction in wall-clock runtime compared to baseline, with all quality metrics (best score, top-k, diversity, hit count) showing parity across all configurations. This confirms that acceleration is achieved without compromising optimization quality. GPU acceleration for diversity estimation (CuPy kernels) is automatically enabled on supported hardware, with fallback to CPU-based RDKit for portability. All results are reported under matched seeds and budgets.

\begin{table}[h!]
\centering
\caption{Matrix speedup and quality parity summary for GuacaMol (20 tasks, 6 seeds, budget 20,000). 
}
\label{tab:matrix_acceleration}
\small
\begin{tabular}{lcccc}
\hline
Config & Mean elapsed (s) & Speedup vs baseline & Time reduction (\%) & Mean best score \\
\hline
baseline\_cpu & 1541.2 & 1.00x & 0.0 & 0.7522 \\
dask\_cpu & 1463.3 & 1.05x & 5.1 & 0.7522 \\
rapids\_v100 & 1349.3 & 1.14x & 12.5 & 0.7522 \\
dask\_rapids\_v100 & 1316.1 & 1.17x & 14.6 & 0.7522 \\
\hline
\end{tabular}
\end{table}

\subsection{Test case 2: multi-objective optimization}

Motivated by the improved performance of SpectralMol on the previous MPO benchmarks, the next step is the evaluation of the proposed framework in a more realistic drug-discovery setting, in which several chemically relevant properties must be optimized simultaneously. For this purpose, the drug-discovery Experiment~2 introduced by Guo et al. \cite{guo2025directly} was reproduced. Among the experiments reported in that study, Experiment~2 was selected as the most relevant benchmark for assessing SpectralMol in a drug-discovery-oriented MPO scenario. In fact, Experiment~1 primarily showed that optimizing docking alone is insufficient, since it tends to produce molecules with unrealistic physicochemical profiles. By contrast, Experiment~2 incorporates QED and SA into a scalar objective function, thus defining a genuine MPO problem that is more closely aligned with practical medicinal-chemistry requirements. The remaining experiments presented by Guo et al. were mainly aimed at analyzing the effect of alternative model initializations, retrosynthesis settings, or specific methodological variations. Although informative within the context of their study, these aspects fall outside the scope of the present comparison, whose objective is to evaluate SpectralMol using a representative and practically relevant drug-discovery benchmark. 

Guo et al. \cite{guo2025directly} used a generative software called Saturn \cite{guo2024saturn} to 
perform the Experiment~2. Saturn is an autoregressive molecular generative model that operates in the space of SMILES strings and is optimized using reinforcement learning. 
The workflow has two distinct learning stages. First, the model is pre-trained on large molecular datasets. 
In this stage, the model learns the syntax of valid molecules.
After pretraining, Saturn enters a reinforcement-learning phase, during which the actual molecule design occurs. In this phase, the model repeatedly samples molecules, evaluates them using a reward function, and updates its parameters to increase the likelihood of generating high-reward molecules.

SpectralMol is compared to Saturn to highlight two key methodological differences.  First the role of training,
SpectralMol is implemented as a population-based evolutionary algorithm and does not require any prior training stage. Candidate molecules are generated and iteratively improved directly from the initial seed population through mutation, crossover, and selection operators, so that the optimization starts immediately once the objective function and the seed pool are specified.
Second, SpectralMol treats these objectives separately and optimizes them jointly using the NSGA-II multi-objective evolutionary algorithm \cite{deb2000fast}, rather than aggregating them into a single scalar function, as implemented in Saturn according to Equation~\ref{eq:rsaqed}. This multi-objective formulation is advantageous because it avoids the need to predefine a fixed trade-off between competing objectives. In a scalarized reward, the relative importance of each term is implicitly determined by the aggregation scheme, which can bias the search toward certain properties and suppress others. By optimizing the objectives independently, NSGA-II maintains a diverse set of Pareto-optimal solutions, allowing for a more comprehensive exploration of the trade-off surface. This is particularly important in molecular design, where objectives such as docking affinity, drug-likeness, and synthesizability are often conflicting, and preserving this diversity enables more flexible downstream selection of candidates.


\begin{equation}
R_{\mathrm{SA\_QED}}(x)
=
\left(
\mathrm{DockingScore}(x)\cdot \mathrm{QED}(x)\cdot \mathrm{SA}(x)
\right)^{1/3}
\in [0,1],
\label{eq:rsaqed}
\end{equation}

In order to preserve the comparability of the benchmark, the SpectralMol experiments were configured to match the published \cite{guo2025directly} constrained-budget setting as closely as possible. In particular, the runs were performed with a fixed oracle-call budget of 1000 evaluations and repeated over 10 independent seeds. The initial population was sampled from the ZINC250k molecular database \cite{irwin2012zinc}, the same resource used for Saturn's pre-training. This shared source improves comparability between the two methods, although it also means that part of the population-level QED/SA behavior reflects initialization from an already drug-like distribution rather than de novo optimization from an unconstrained prior.

Docking simulations were performed with QuickVina2-GPU 2.1 against the ATP-dependent Clp protease proteolytic subunit (ClpP, PDB ID: 6U0J), using a fixed search box centered on the known active site. 
The receptor was prepared once as a rigid PDBQT structure, with non-protein species (co-crystallized ligands, solvent, and ions) removed and Gasteiger charges assigned. Ligands were prepared following the Saturn (Guo et al.~\cite{guo2025directly}) protocol. Docking used the same QuickVina2-GPU settings across runs (thread/exhaustiveness fixed; heuristic search depth; num\_modes = 1; fixed random seed), and the best (lowest-energy) pose score was retained per molecule (or per prepared variant, taking the minimum across variants). Duplicate molecules were identified and served from cache (not re-docked). 
Oracle-call accounting counted newly evaluated  molecules, including docking-failed molecules assigned the failure score, while cached duplicates were not counted as new oracle evaluations. Saturn results were taken directly from Guo et al.~\cite{guo2025directly}, where the same docking engine, protein target, and search-box specification were used; thus, the comparison is intended to isolate differences in optimization strategy rather than differences in oracle definition. Table \ref{tab:clpp_reproducibility} reports the settings used for the experiment.

\begin{table}[h!]
\centering
\caption{Reproducibility summary for the ClpP docking benchmark used in Test case 2.}
\label{tab:clpp_reproducibility}
\small
\begin{tabular}{ll}
\hline
Target protein & ATP-dependent Clp protease proteolytic subunit (ClpP) \\
PDB ID & 6U0J \\
Docking engine & QuickVina2 GPU 2.1 \\
Search-box center & $(27.0,\ 27.5,\ 24.0)$ \\
Search-box size & $(18,\ 18,\ 18)$~\AA \\
Ligand preparation & Guo et al. \cite{guo2025directly}  \\
Oracle budget & 1000 evaluations per run \\
Replicates & 10 independent seeds \\
Comparator source & Saturn values from Guo et al. \cite{guo2025directly} \\
\hline
\end{tabular}
\end{table}

Table \ref{Saturn} provides a direct comparison between published Saturn results for Experiment~2 performed by Guo et al.~\cite{guo2025directly} using the Saturn generative algorithm~\cite{guo2024saturn} and the same experiment recreated with SpectralMol. Results are reported for two docking-threshold cases: molecules passing a threshold of $< -9$ kcal/mol, and a more stringent threshold of $< -10$ kcal/mol. Under the same oracle, both methods generate molecules that satisfy the imposed docking criteria while maintaining reasonable physicochemical profiles. 
For SpectralMol, a seed is denoted as successful at a given docking threshold if it produced at least one molecule satisfying that threshold. SpectralMol means and standard deviations are computed across all ten seeds, assigning zero Modes and zero Yield to seeds with no threshold-passing molecules. 
Mean and standard deviation of QED, SA, and molecular weight are computed over threshold-passing molecules only; therefore, molecular property statistics are defined only for successful seeds. 
At $< -9$ kcal/mol, SpectralMol shows a promising improvement in both diversity and hit count, with substantially higher Modes and Yield (76.3 $\pm$ 44.5 and 133.2 $\pm$ 55.8) than Saturn (38.0 $\pm$ 13.0 and 83.0 $\pm$ 33.0). SpectralMol also retains slightly higher QED (0.82 vs.\ 0.80), but with a higher (worse) SA score (2.79 vs.\ 2.10), indicating a diversity/hit-rate gain with somewhat reduced synthetic simplicity. At the stricter $< -10$ kcal/mol threshold, SpectralMol again reports higher mean Modes and Yield among reported runs (5.9 $\pm$ 8.5 and 6.5 $\pm$ 9.2) than Saturn (3.0 $\pm$ 1.0 and 4.0 $\pm$ 2.0), but with larger variance and fewer successful replicates (7 vs.\ 9), suggesting lower consistency under the most stringent constraint. Overall, the table indicates that SpectralMol improves exploration and hit generation, while Saturn remains more stable at the hardest threshold.

The increase in diversity and yield with SpectralMol does not substantially degrade average drug-likeness, as QED (slightly better for SpectralMol) and molecular weight are similar across methods, but it is accompanied by higher (worse) SA scores. SpectralMol's broader exploration appears to preserve favorable property ranges while shifting part of the hit set toward compounds that are predicted to be less synthetically accessible.

\begin{table}[h!]
\centering
\setlength{\tabcolsep}{4pt}
\caption{
Saturn benchmark and SpectralMol comparison at 1000 oracle calls.
For each method/threshold, the number in parentheses is the number of successful replicates (out of 10 seeds), defined as seeds with at least one docking hit at that threshold.
For Saturn$^*$, values are reported from Guo et al.~\cite{guo2025directly}.
For SpectralMol, values are reported as mean $\pm$ standard deviation across all the seeds under the same thresholding rule.
\emph{Modes} is the number of unique scaffolds among threshold-passing molecules, and \emph{Yield} is the number of threshold-passing molecules.
}
\label{Saturn}
\resizebox{\textwidth}{!}{
\begin{tabular}{lllll}
\hline
Method (successful replicates) & Modes (Yield) & QED & SA score & MolWt \\
\hline
\multicolumn{5}{l}{\textbf{Docking score $< -9$ kcal/mol}} \\
Saturn$^*$ (10) & 38.0 $\pm$ 13.0 (83.0 $\pm$ 33.0) & 0.80 $\pm$ 0.05 & 2.10 $\pm$ 0.10 & 343.3 $\pm$ 6.2 \\
SpectralMol (10) & 76.3 $\pm$ 44.5 (133.2 $\pm$ 55.8) & 0.82 $\pm$ 0.03 & 2.79 $\pm$ 0.39 & 341.3 $\pm$ 33.6 \\
\hline
\multicolumn{5}{l}{\textbf{Docking score $< -10$ kcal/mol}} \\
Saturn$^*$ (9) & 3.0 $\pm$ 1.0 (4.0 $\pm$ 2.0) & 0.72 $\pm$ 0.16 & 2.26 $\pm$ 0.16 & 379.9 $\pm$ 18.8 \\
SpectralMol (7) & 5.9 $\pm$ 8.5 (6.5 $\pm$ 9.2) & 0.79 $\pm$ 0.07 & 2.93 $\pm$ 0.32 & 365.7 $\pm$ 29.9 \\
\hline
\multicolumn{5}{l}{$^*$ Guo et al.~\cite{guo2025directly}.} \\
\end{tabular}
}
\end{table}

\begin{figure}[h]
    \centering
    \begin{subfigure}{0.45\textwidth}
        \centering
        \includegraphics[width=\linewidth]{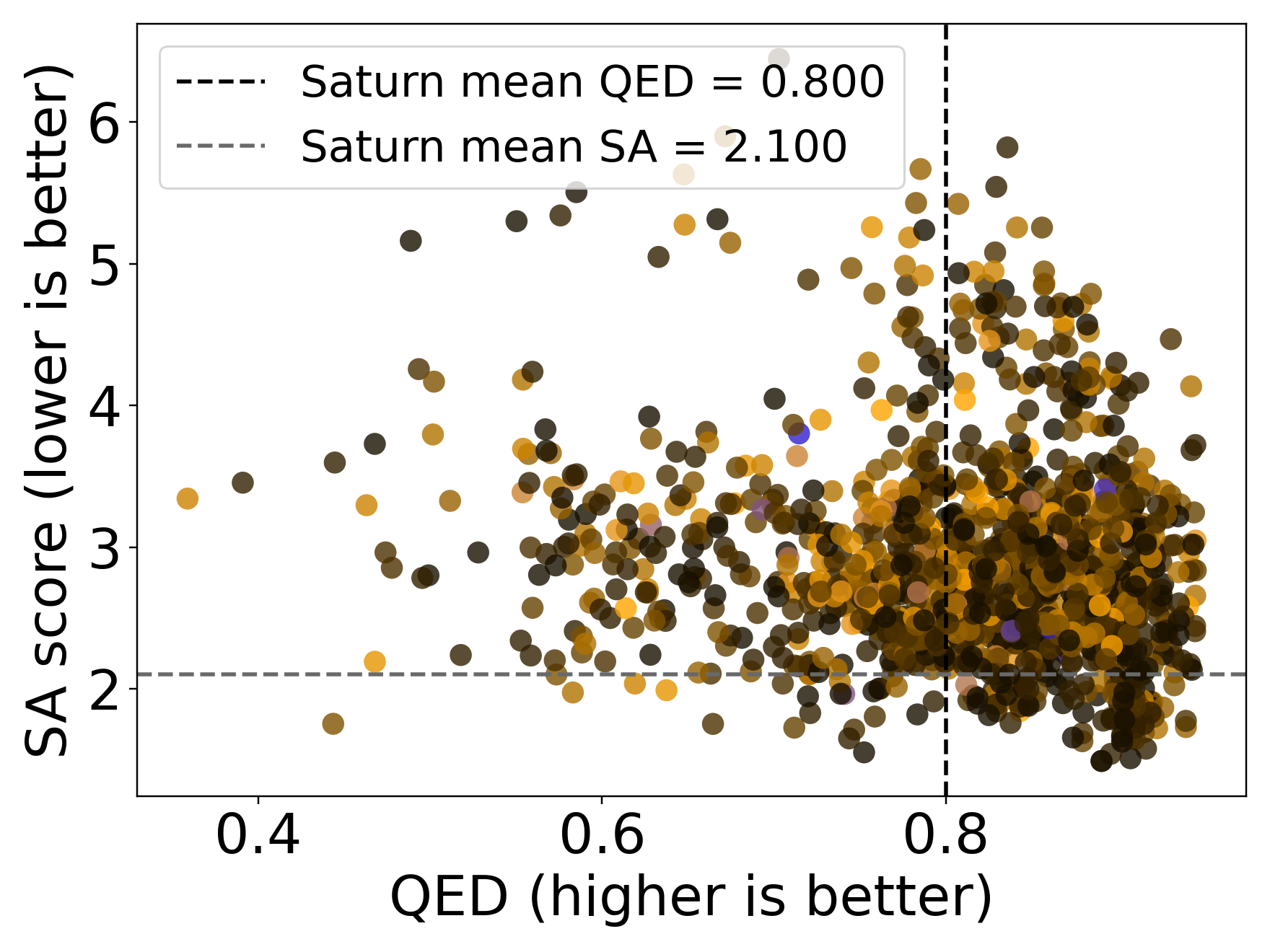}
        \caption{Docking score $<-9$ kcal/mol (n=1332)}
        \label{fig:fig1}
    \end{subfigure}
    \begin{subfigure}{0.45\textwidth}
        \centering
        \includegraphics[width=\linewidth]{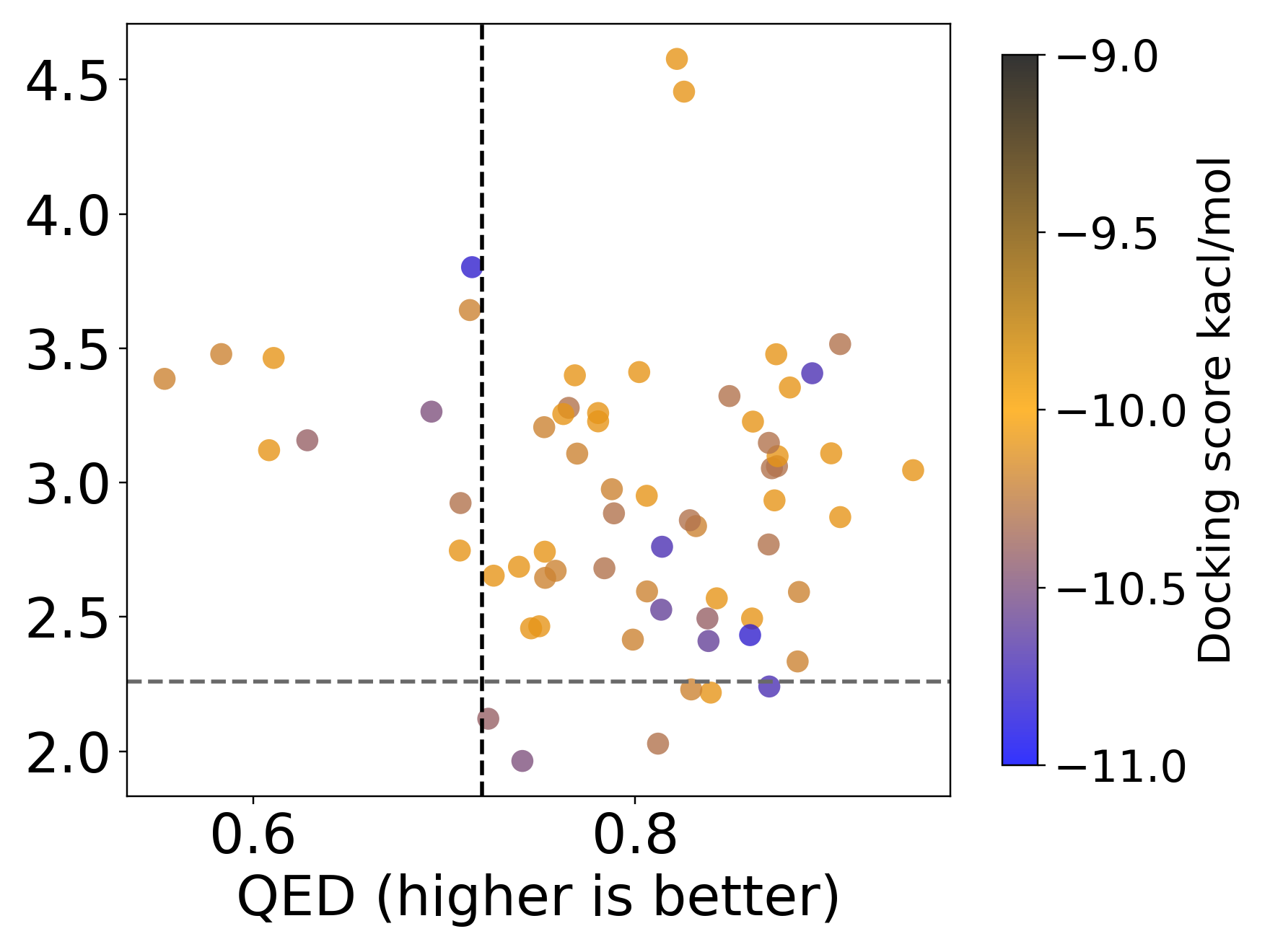}
        \caption{Docking score $<-10$ kcal/mol (n=65)}
        \label{fig:fig2}
    \end{subfigure}
    \caption{Distribution of molecules (n) in QED–SA space passing docking score thresholds. Points are color-coded by docking score, with higher QED (right) and lower SA (down) indicating more drug-like and synthetically accessible molecules. Dashed lines denote average QED and SA values for Saturn \cite{guo2025directly}. }
    \label{fig:combined}
\end{figure}

Figure \ref{fig:combined}(a) shows the molecular distribution in QED–SA space for all compounds passing the docking threshold of $-9$\,kcal/mol. The scatter plot reveals a dense clustering of molecules in the region characterized by high QED values ($>$0.8) and moderate SA scores (2 to 3).
On average, QED exceeds Saturn’s benchmark, whereas SA performs slightly worse. However, the distribution highlights a substantial subset of molecules occupying the desirable region of higher QED (right) and lower SA (down). Notably, 109 molecules surpass Saturn’s average metrics on both criteria, representing approximately 8\% of the total dataset.

In the more stringent regime, SpectralMol shows a higher mean QED than Saturn, whereas both methods shift toward higher SA values, indicating reduced synthetic accessibility among the strictest docking hits.
Figure~\ref{fig:combined}(b) provides further insight into this regime. 
Compared to the $< -9$ kcal/mol case, the distribution is significantly sparser, indicating the difficulty of identifying high-affinity candidates. 
However, the remaining molecules are still predominantly located in regions of relatively high QED. The dispersion in SA scores is more pronounced, suggesting that achieving very low docking scores may require exploring more synthetically complex regions of the chemical space. Six molecules, corresponding to roughly 9\% of population passing the strict-threshold, outperform Saturn's average metrics on both QED and SA.

In summary, At the $-9$\,kcal/mol threshold, SpectralMol yields higher mean scaffold diversity and hit count than the published Saturn results. At the stricter $-10$\,kcal/mol threshold, SpectralMol also has higher conditional mean modes and yield, but with lower replicate success and high variance.
These results suggest that SpectralMol can effectively explore chemical space, consistently yielding higher diversity and hit rates than Saturn while preserving good physicochemical properties, at the cost of increased synthesis complexity.

\begin{figure}[htbp]
    \centering
    \begin{subfigure}[b]{0.32\textwidth}
        \centering
        \includegraphics[width=\textwidth]{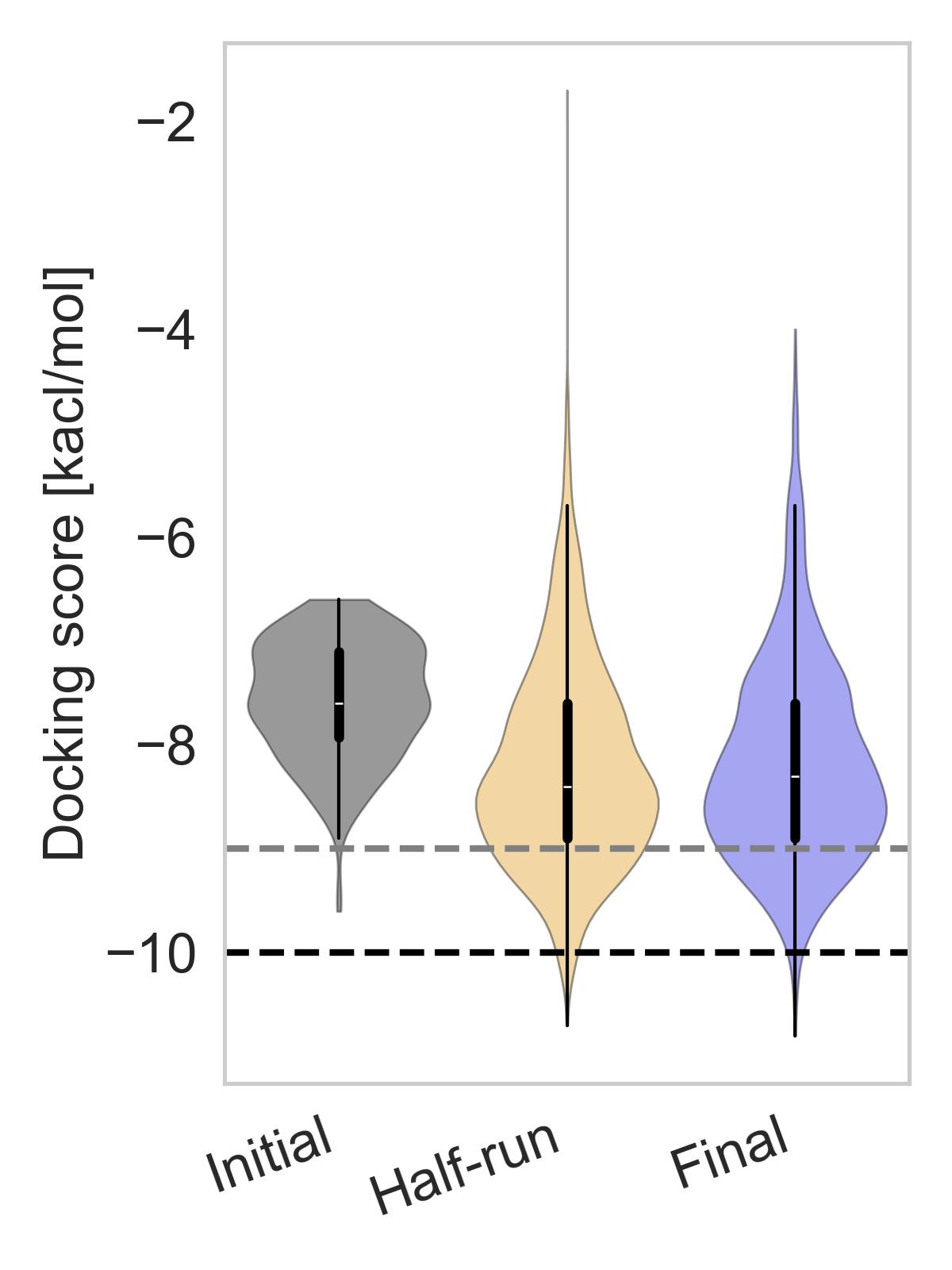}
        \caption{Docking score evolution.}
    \end{subfigure}
    \hfill
    \begin{subfigure}[b]{0.32\textwidth}
        \centering
        \includegraphics[width=\textwidth]{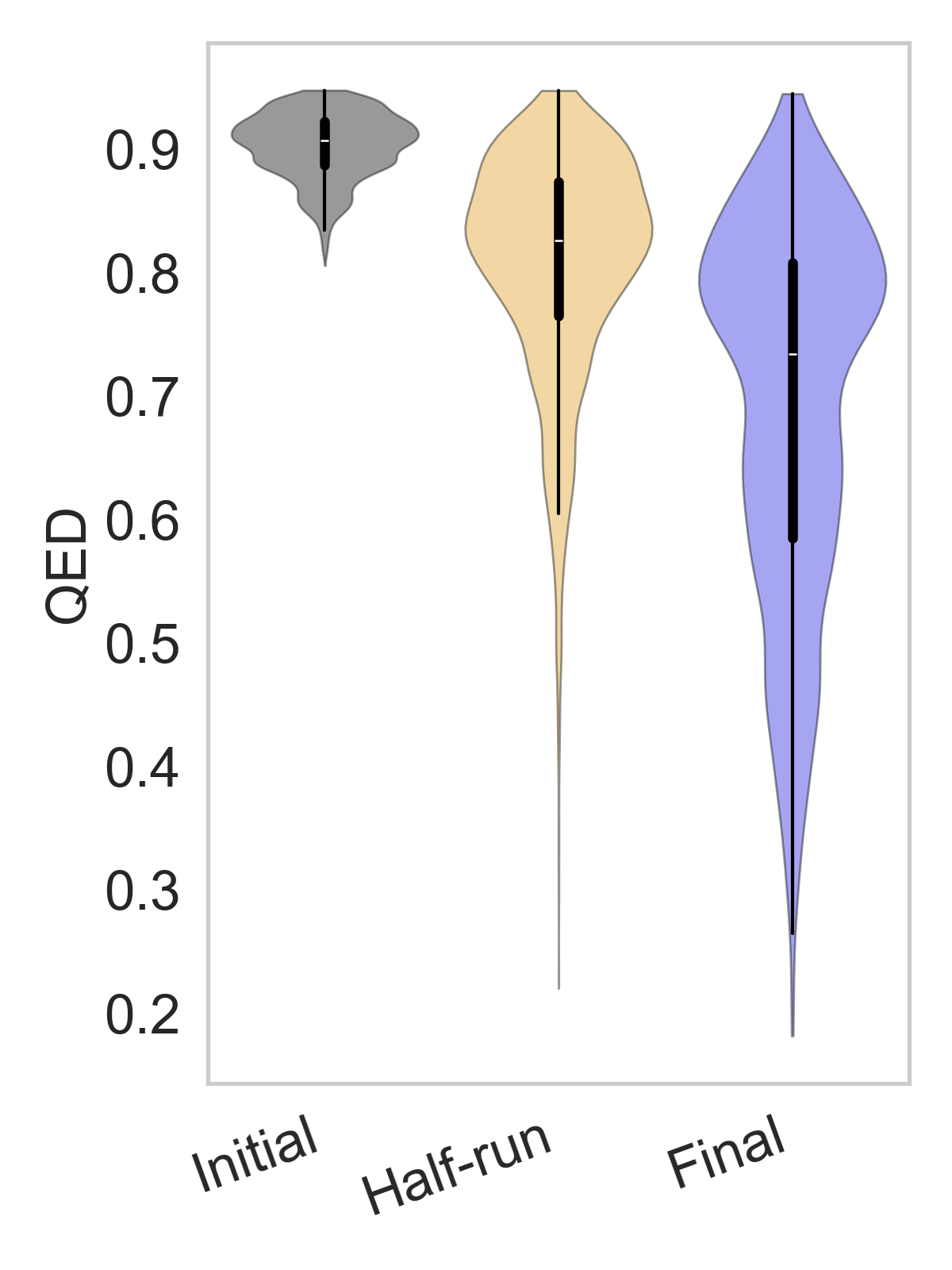}
        \caption{QED evolution.}
    \end{subfigure}
    \hfill
    \begin{subfigure}[b]{0.32\textwidth}
        \centering
        \includegraphics[width=\textwidth]{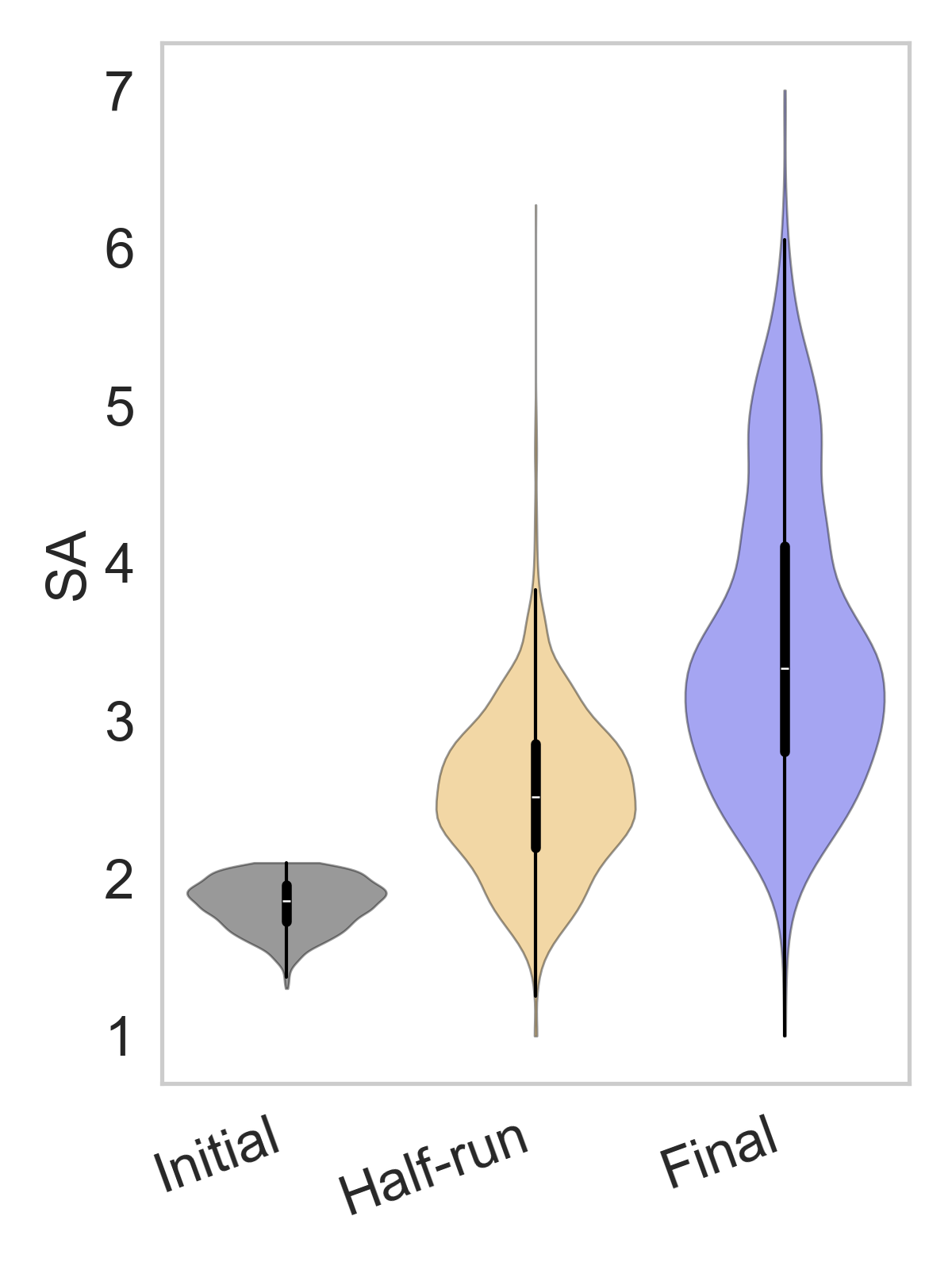}
        \caption{SA evolution.}
    \end{subfigure}

    \vspace{0.5cm}

    \begin{subfigure}[b]{0.32\textwidth}
        \centering
        \includegraphics[width=\textwidth]{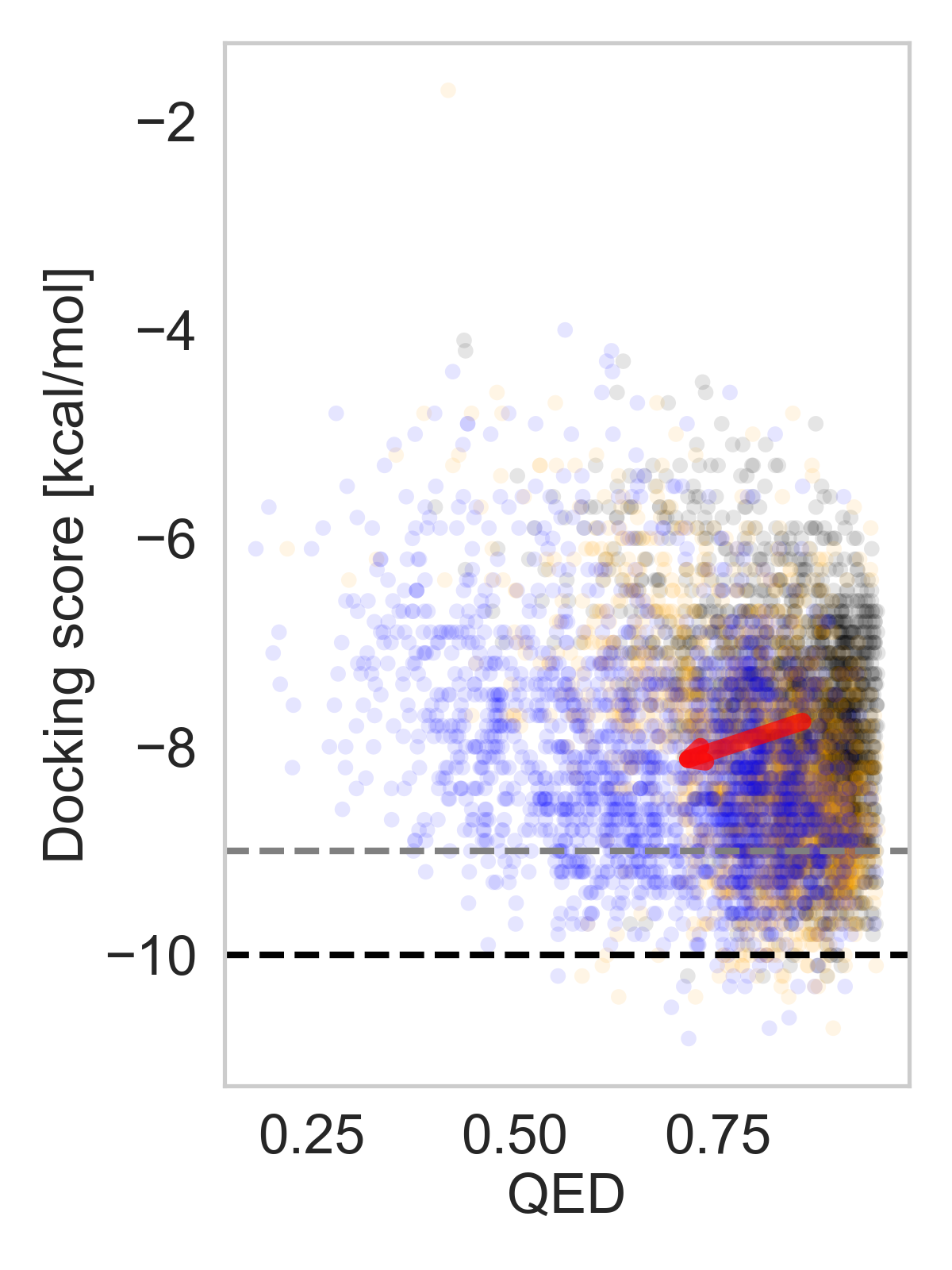}
        \caption{Distributions in QED-docking space.}
    \end{subfigure}
    \hfill
    \begin{subfigure}[b]{0.32\textwidth}
        \centering
        \includegraphics[width=\textwidth]{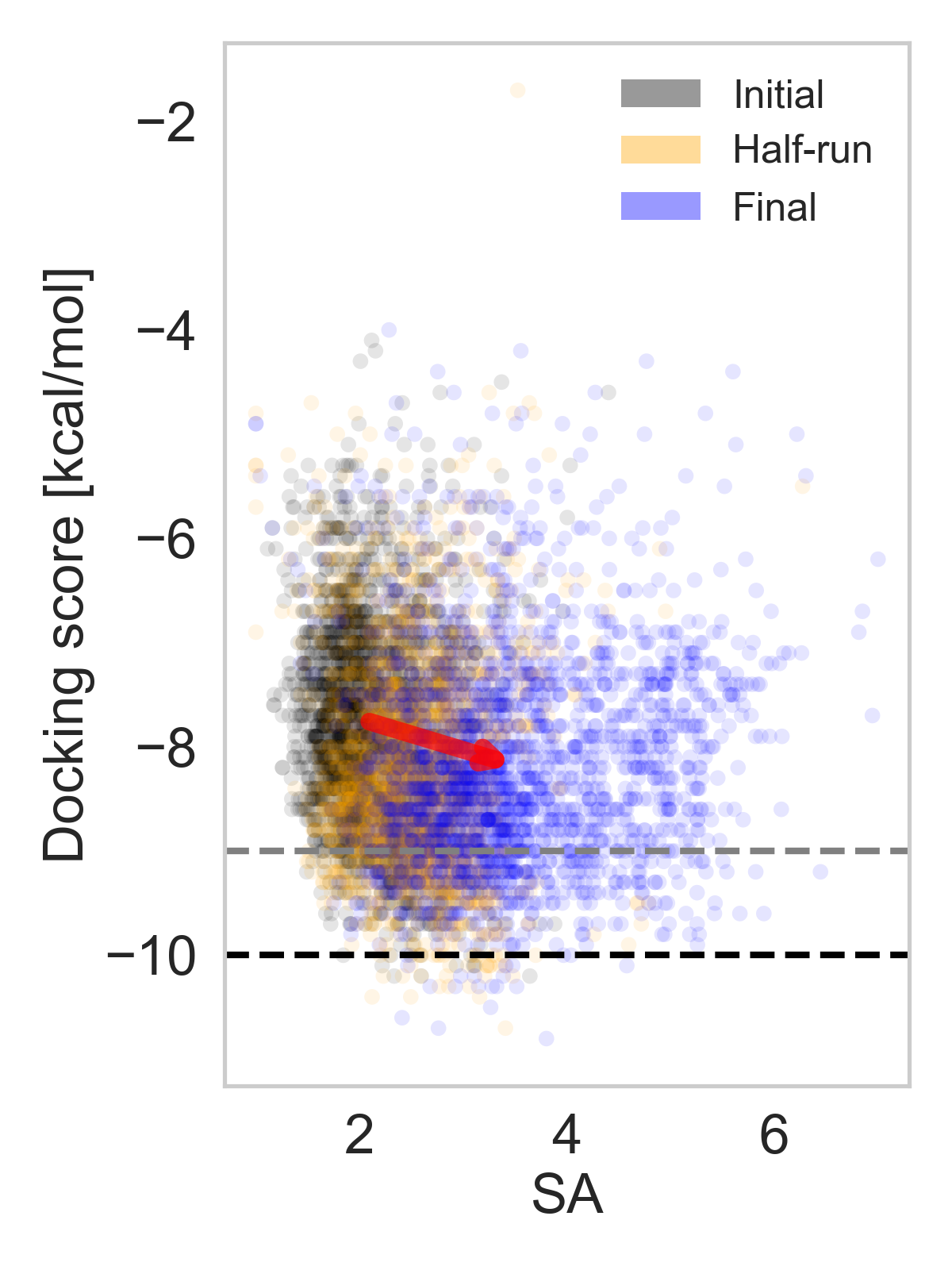}
        \caption{Distributions in SA-docking space.}
    \end{subfigure}
    \hfill
    \begin{subfigure}[b]{0.32\textwidth}
        \centering
        \includegraphics[width=\textwidth]{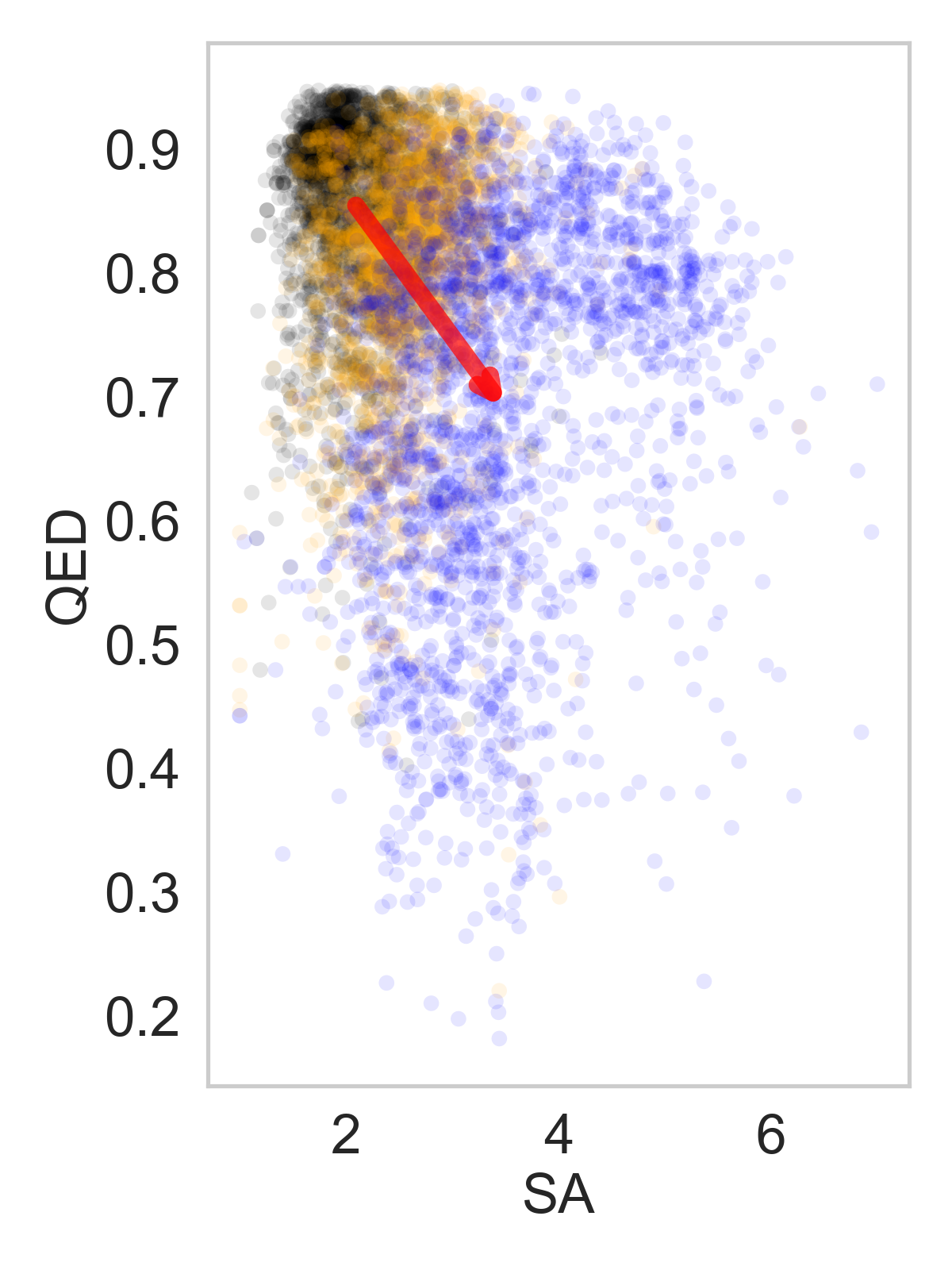}
        \caption{Distributions in QED-SA space.}
    \end{subfigure}

    \caption{SpectralMol optimization evolution comparing initial, half-run, and final population of molecules. The top row shows the density distribution for the objective properties. The bottom row shows projections of the same objectives in 2D space, highlighting the structure of the explored chemical space.}
    \label{optimization_evolution}
\end{figure}

\paragraph{Optimization evolution and pareto front analysis}
The objective of the following analysis is to evaluate the evolution of the docking score, QED, and SA throughout the optimization run. To this end, Figure \ref{optimization_evolution} provides an overview of the optimization dynamics by comparing the distributions of these properties in the initial population, at half of the allowed oracle calls, and in the final population. Each distribution is evaluated over 2,560 molecules, corresponding to a population of 256 molecules for each of the ten seeds. The dotted lines indicate the docking score thresholds used in the comparison with Saturn.

Figure~\ref{optimization_evolution} shows, in the top row, the evolution of the density distributions. A clear downward shift in the docking score distribution (Figure~\ref{optimization_evolution} (a)) is observed, indicating that the evolutionary process effectively drives the population toward better predicted binding affinity. In contrast, the QED distribution shifts toward lower values (Figure~\ref{optimization_evolution} (b)), while the raw SA distribution shifts toward higher values (Figure~\ref{optimization_evolution} (c)), both corresponding to less favorable molecular profiles.
This behavior can be attributed to two main factors. First, the initial population is sampled from a drug-like ZINC250k-derived seed pool and is therefore already biased toward high QED and low raw SA values. Second, the multi-objective search introduces a trade-off: improving docking affinity often requires exploring broader regions of chemical space, which can reduce drug-likeness and increase synthetic difficulty. Consequently, the evolutionary process expands toward regions that improve predicted binding performance while partially sacrificing QED and synthetic accessibility.

The bottom row of Figure~\ref{optimization_evolution} further clarifies these relationships through projections in two-dimensional objective spaces. The red arrow shows the evolution of the mean of the distribution in each projection, highlighting the overall direction of population-level directions.
Figure~\ref{optimization_evolution} (d) reveals that the initial population (black) is concentrated in a relatively narrow region characterized by high QED and moderately good docking scores. 
As the evolutionary process progresses, the final population (blue) shifts toward lower docking scores, indicating improved binding performance, while exhibiting a reduction in average QED.
In addition to this shift, the final population is noticeably more dispersed, spanning a wider range of both QED and docking values. This increased spread reflects the exploratory nature of the evolutionary algorithm: while high-performing individuals are preserved through selection, variation operators such as mutation and crossover continuously introduce diversity, enabling the population to explore multiple regions of the objective space. As a result, the algorithm does not converge to a single narrow optimum but instead maintains a heterogeneous set of candidates with different trade-offs between objectives.
Similar discussion can be extended to the docking–SA case, reported in Figure~\ref{optimization_evolution} (e).
Figure~\ref{optimization_evolution} (f) presents the QED–SA projection. The initial population is clustered in the region of high QED and low SA, consistent with the model initialization. The final population becomes significantly more spread toward lower QED and higher SA values. 

In summary, the evolution from the initial to the final population is characterized by a shift toward improved docking performance accompanied by a broadening of the explored objective space. While the initial population is tightly clustered in a region of high QED and low SA, the final population becomes more dispersed, extending toward lower QED and higher SA values. This indicates that the optimization process does not simply refine the initial distribution but actively explores new regions of chemical space, generating a diverse set of candidates with different trade-offs between stronger binding, drug-likeness, and synthetic accessibility.

Finally, the Pareto front and crowding distance were evaluated to assess the number of non-dominated molecules and their diversity in the considered optimization space. 
The first non-dominated front ($F_1$) is defined as the subset of the final population $P$ for which no molecule is Pareto-dominated by any other member of $P$, where $y \prec x$ if $y$ is at least as good as $x$ in all objectives (higher QED, lower SA, and lower docking score) with at least one strict improvement.
\[
F_1=\{x\in P:\nexists\, y\in P \text{ such that } y \prec x\},
\]

\begin{figure}[htbp]
    \centering
    \begin{subfigure}[b]{0.32\textwidth}
        \centering
        \includegraphics[width=\textwidth]{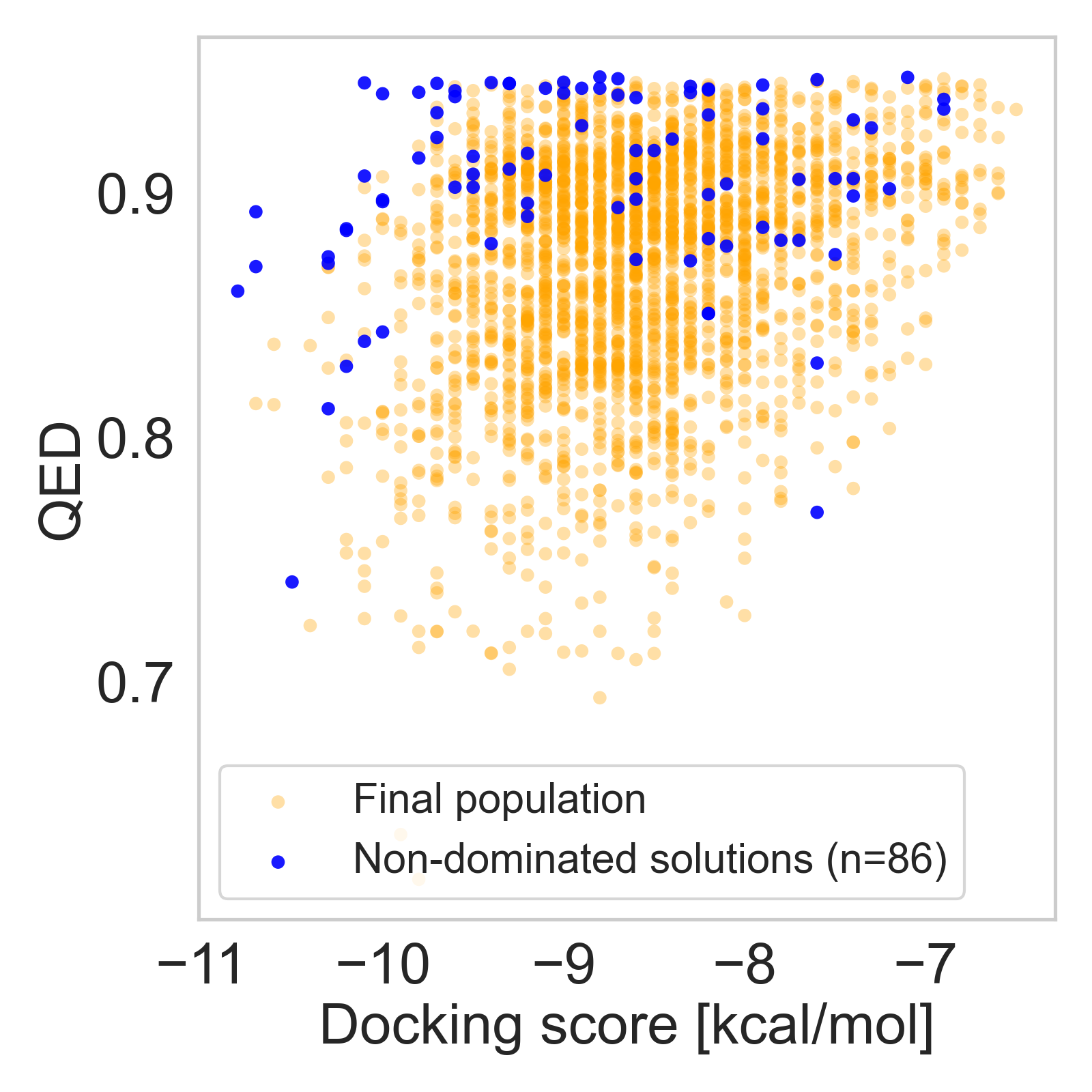}
        \caption{QED-docking space.}
    \end{subfigure}
    \hfill
    \begin{subfigure}[b]{0.32\textwidth}
        \centering
        \includegraphics[width=\textwidth]{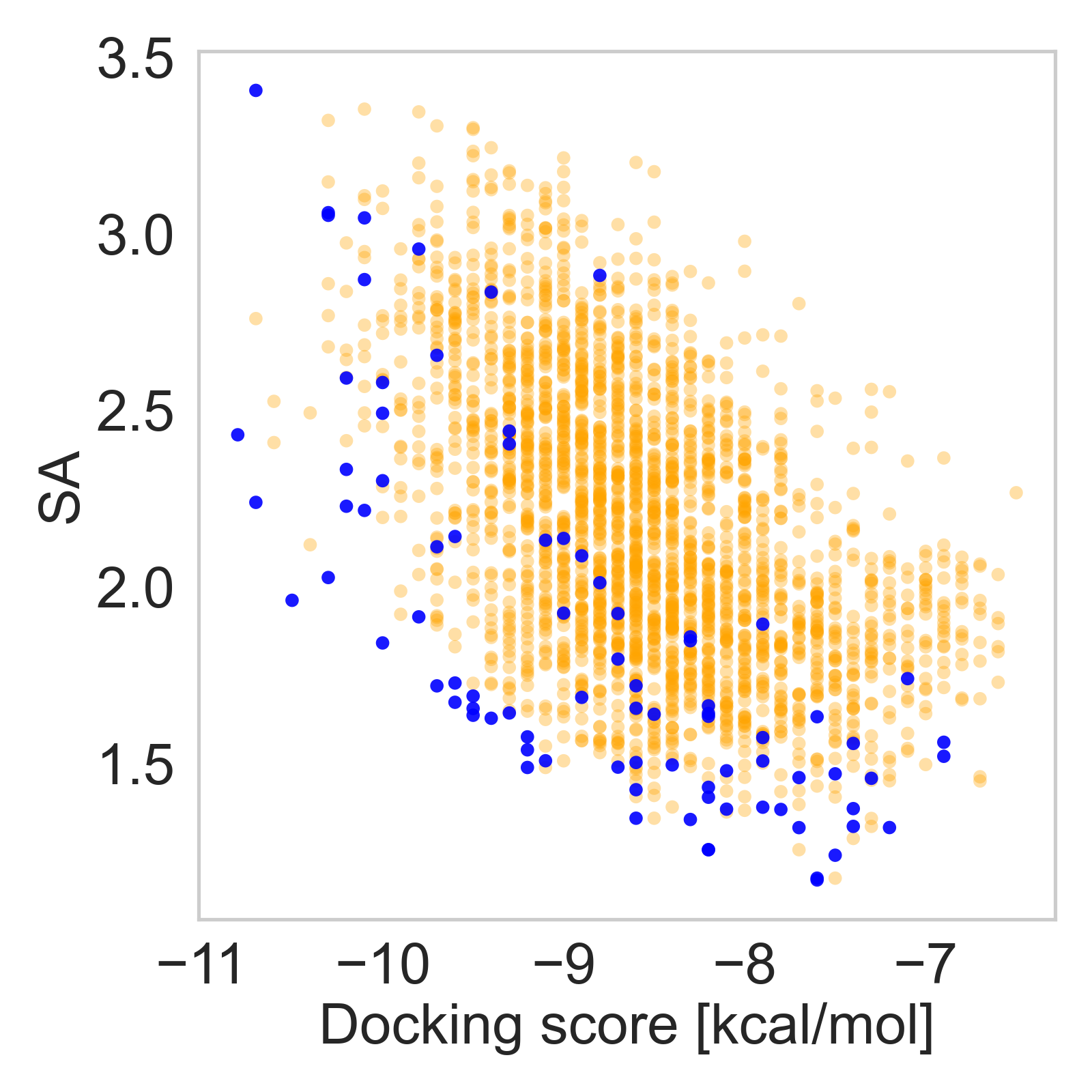}
        \caption{SA-docking space.}
    \end{subfigure}
    \hfill
    \begin{subfigure}[b]{0.32\textwidth}
        \centering
        \includegraphics[width=\textwidth]{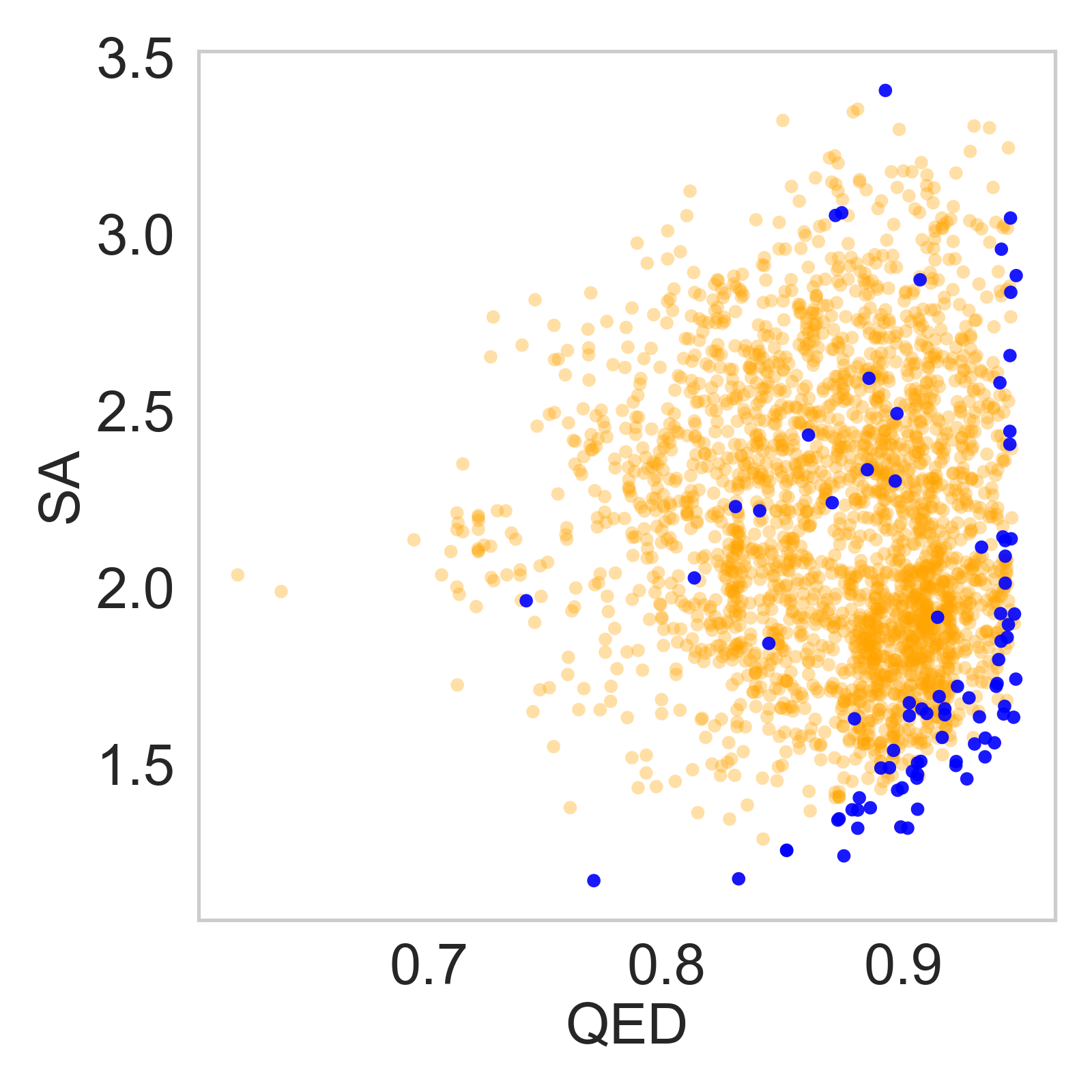}
        \caption{QED-SA space.}
    \end{subfigure}

    \caption{First non-dominated front $F_1$ (blue, $n=86$) plotted against the full final population $P$ (orange, $n=2{,}560$ molecules pooled across 10 seeds, 256 molecules per seed) in three two-dimensional projections of the QED--SA--docking objective space: (a) QED versus docking score, (b) SA versus docking score, and (c) QED versus SA. The non-dominated set is enriched in regions combining favourable docking, high QED, and low SA, while occupying only $3.36\%$ of the final population. $F_1$ molecules show $\mathrm{QED}=0.905\pm0.040$, $\mathrm{SA}=1.86\pm0.52$, and a mean docking score of $-8.85\pm1.01$\,kcal/mol; only $44.2\%$ of $F_1$ satisfies docking $<-9$\,kcal/mol and $15.1\%$ satisfies docking $<-10$\,kcal/mol, indicating that NSGA-II preserves chemically balanced trade-offs rather than collapsing the search onto a single objective.}
    \label{Pareto_front}
\end{figure}

In the SpectralMol final population, comprising 2,560 molecules with 256 molecules per seed across 10 seeds, 86 molecules were identified in $F_1$, representing 3.36\% of the population. 
Figure~\ref{Pareto_front} highlights the $F_1$ molecules in blue against the final population $P$ in orange. 
Figure~\ref{Pareto_front} (a) shows the QED--docking space, where $F_1$ is enriched in molecules combining favorable docking scores with high drug-likeness. Figure~\ref{Pareto_front} (b) presents the SA--docking space, indicating that the non-dominated solutions maintain competitive docking while tending toward lower raw SA values, consistent with improved synthetic accessibility. Figure~\ref{Pareto_front} (c) shows the QED--SA space, where $F_1$ occupies a region characterized by high QED and low raw SA relative to the full population.
Overall, $F_1$ is characterized by high drug-likeness and favorable synthetic accessibility, with QED $=0.905\pm0.040$ and raw SA $=1.86\pm0.52$, while maintaining competitive docking performance with a mean docking score of $-8.85\pm1.01$. Importantly, the non-dominated set does not maximize docking alone: only 44.2\% of $F_1$ satisfies docking $<-9$, and 15.1\% satisfies docking $<-10$. This indicates that NSGA-II preserves chemically balanced trade-offs rather than collapsing onto a single objective.

Within $F_1$, diversity is quantified using the crowding distance, which measures the sparsity of each solution in objective space by summing normalized nearest-neighbor gaps across objectives. For a front with $J$ objectives, the crowding distance of a molecule $x_i$ is computed by sorting the front separately according to each objective and summing the normalized absolute distance between neighboring objective values. Let $f_j(\cdot)$ denote the value of objective $j$, and let $x_{i-1}^{(j)}$ and $x_{i+1}^{(j)}$ be the neighboring solutions of $x_i$ after sorting the front by objective $j$. The corresponding crowding distance is defined as:
\[
\mathrm{CD}(x_i)=\sum_{j=1}^{J}
\frac{\left|f_j(x_{i+1}^{(j)})-f_j(x_{i-1}^{(j)})\right|}
{f_j^{\max}-f_j^{\min}},
\]
where $f_j^{\max}$ and $f_j^{\min}$ are the maximum and minimum values of objective $j$ within the front, respectively. Objectives with zero range, i.e., $f_j^{\max}=f_j^{\min}$, are omitted from the sum to avoid division by zero. Boundary solutions are conventionally assigned $\mathrm{CD}=\infty$ to preserve extreme points of the trade-off surface.
Thus, larger crowding distance values correspond to $F_1$ molecules occupying less populated regions of objective space, whereas smaller values indicate molecules located in denser regions and therefore greater local redundancy. In $F_1$, the median crowding distance is 0.0536, with a 90th percentile of 0.1114.
Among docking hits, median crowding increases for stricter affinity thresholds (0.0499 at docking $<-9$ vs 0.1006 at docking $<-10$), consistent with distributions reported in Figure \ref{fig:combined}. 

\begin{figure}
    \centering
    \includegraphics[width=0.33\linewidth]{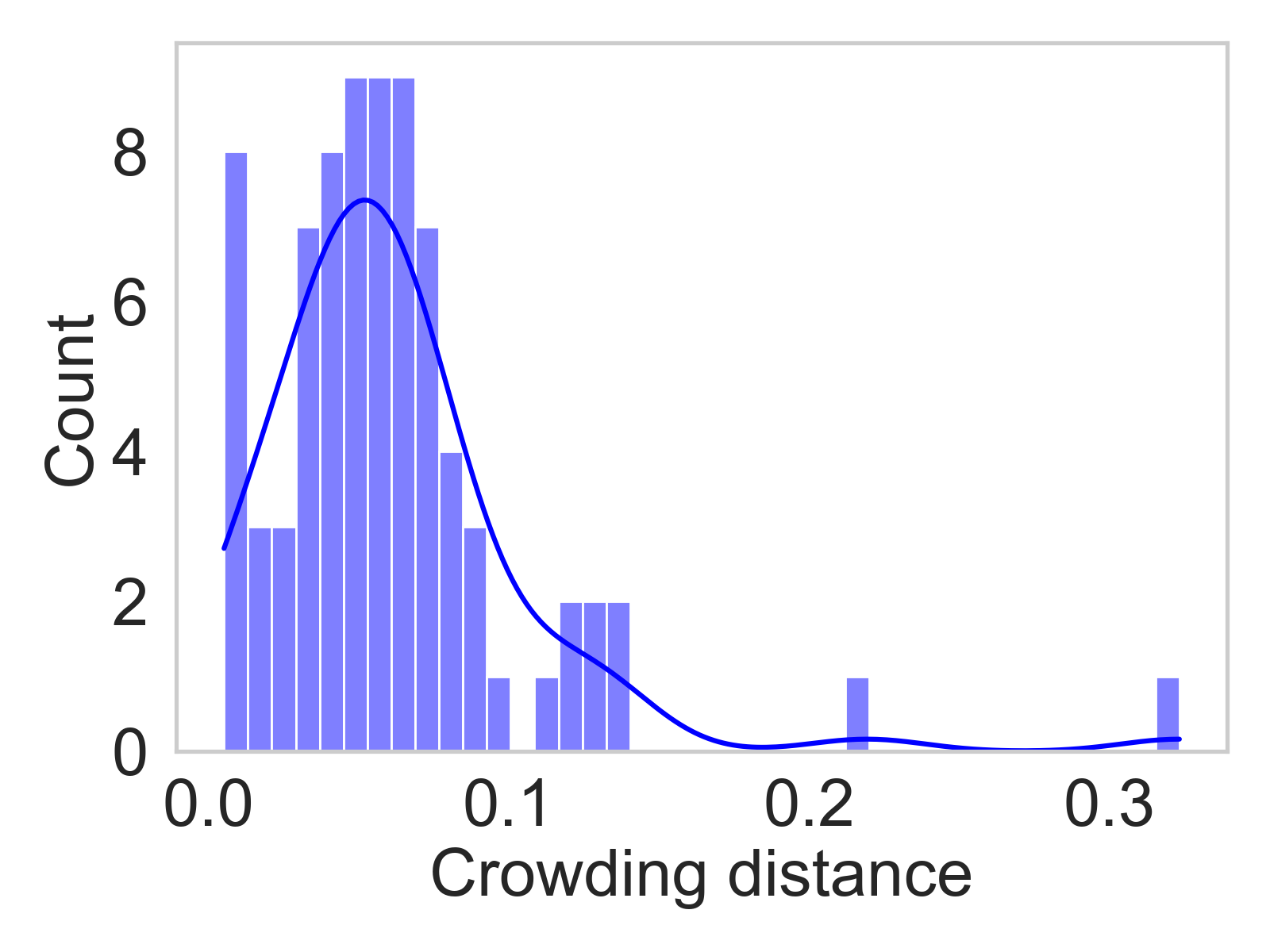}
    \caption{Distribution of crowding distances for the $86$ molecules of the first non-dominated front $F_1$, with kernel-density estimate overlaid (solid line). The histogram is concentrated at low crowding-distance values (median $0.0536$, $90$th percentile $0.1114$), indicating that a substantial fraction of $F_1$ occupies locally dense regions of objective space. The clear right tail identifies a smaller subset of molecules with markedly larger crowding distances, corresponding to isolated and therefore more diverse Pareto-optimal solutions. Boundary solutions, conventionally assigned $\mathrm{CD}=\infty$, are excluded from the histogram.}
    \label{crowd distance}
\end{figure}

Figure \ref{crowd distance} shows the distribution of crowding distances for the $F_1$ molecules. Most of the molecules are concentrated at relatively low crowding-distance values, indicating that a substantial fraction of $F_1$ occupies locally dense regions of objective space. At the same time, the distribution exhibits a clear right tail, demonstrating the presence of a smaller subset of molecules with markedly larger crowding distances that occupy sparser regions of the front and thus contribute to increase its diversity. 

\paragraph{Molecular analysis}

\begin{figure}
    \centering
    \includegraphics[width=\linewidth]{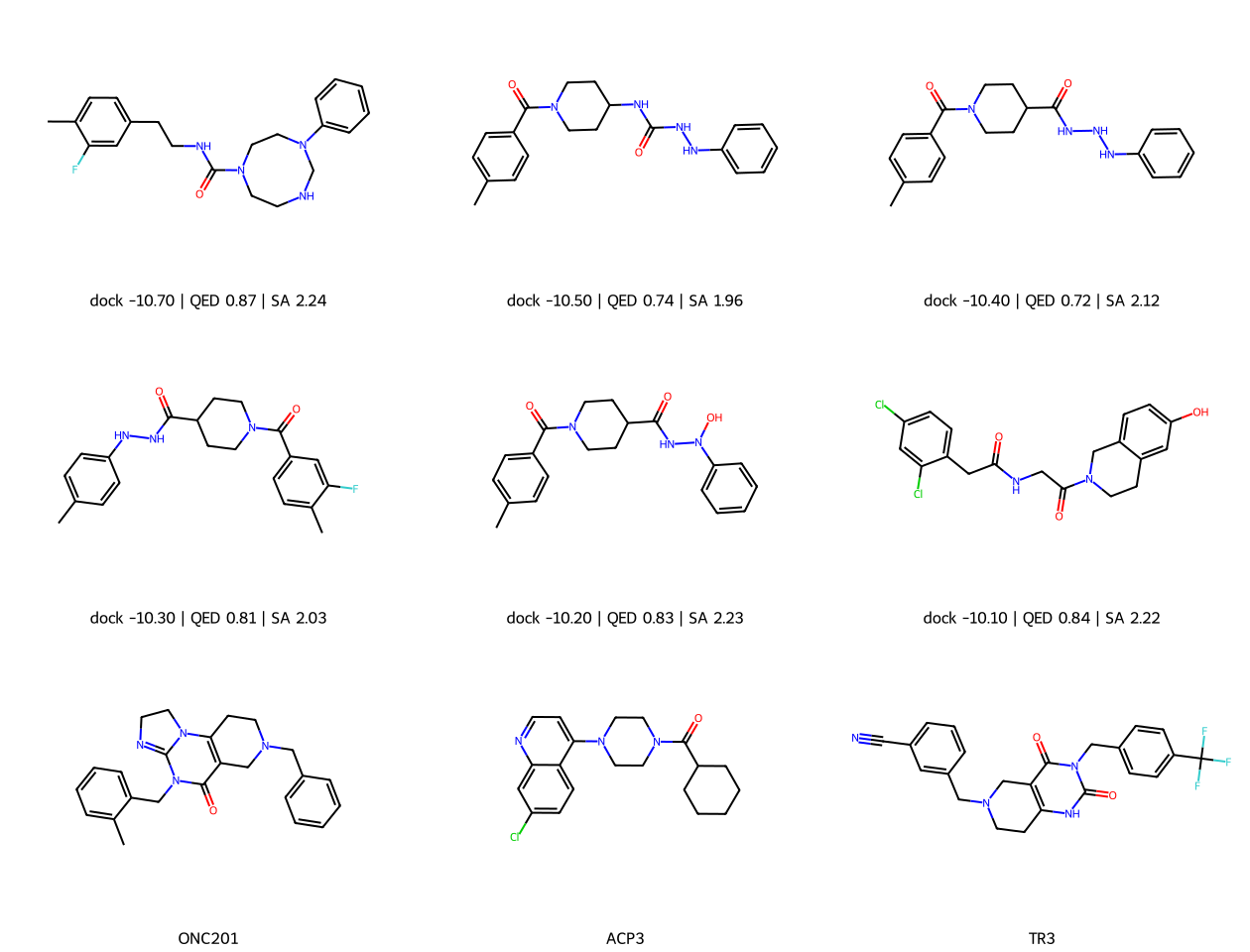}
    \caption{First and second rows: SpectralMol-generated candidates passing docking score $<-10$\,kcal/mol, QED $>0.72$, and raw SA $<2.26$. Last row: reference ClpP-targeting compounds reported in the literature~\cite{miciaccia2025onc201,ye2017development}.}
    \label{molecules}
\end{figure}

To evaluate the chemical relevance of the molecules generated by SpectralMol, we compared the candidates passing the selection criteria (docking score $< -10$ kcal/mol, QED $> 0.72$, and SA $< 2.26$) with known ClpP-targeting compounds.
Figure \ref{molecules} presents the SpectralMol-generated molecules and, in the last row, three real molecules used for the activation of the ClpP protease \cite{miciaccia2025onc201,ye2017development}.
The generated molecules (first and second rows) exhibit structural features consistent with known ClpP ligands. In particular, both generated and reference compounds share common architectural motifs, such as terminal aromatic rings, carbonyl linkers, and central nitrogen-containing structures.

\begin{figure}
    \centering
    \includegraphics[width=0.8\linewidth]{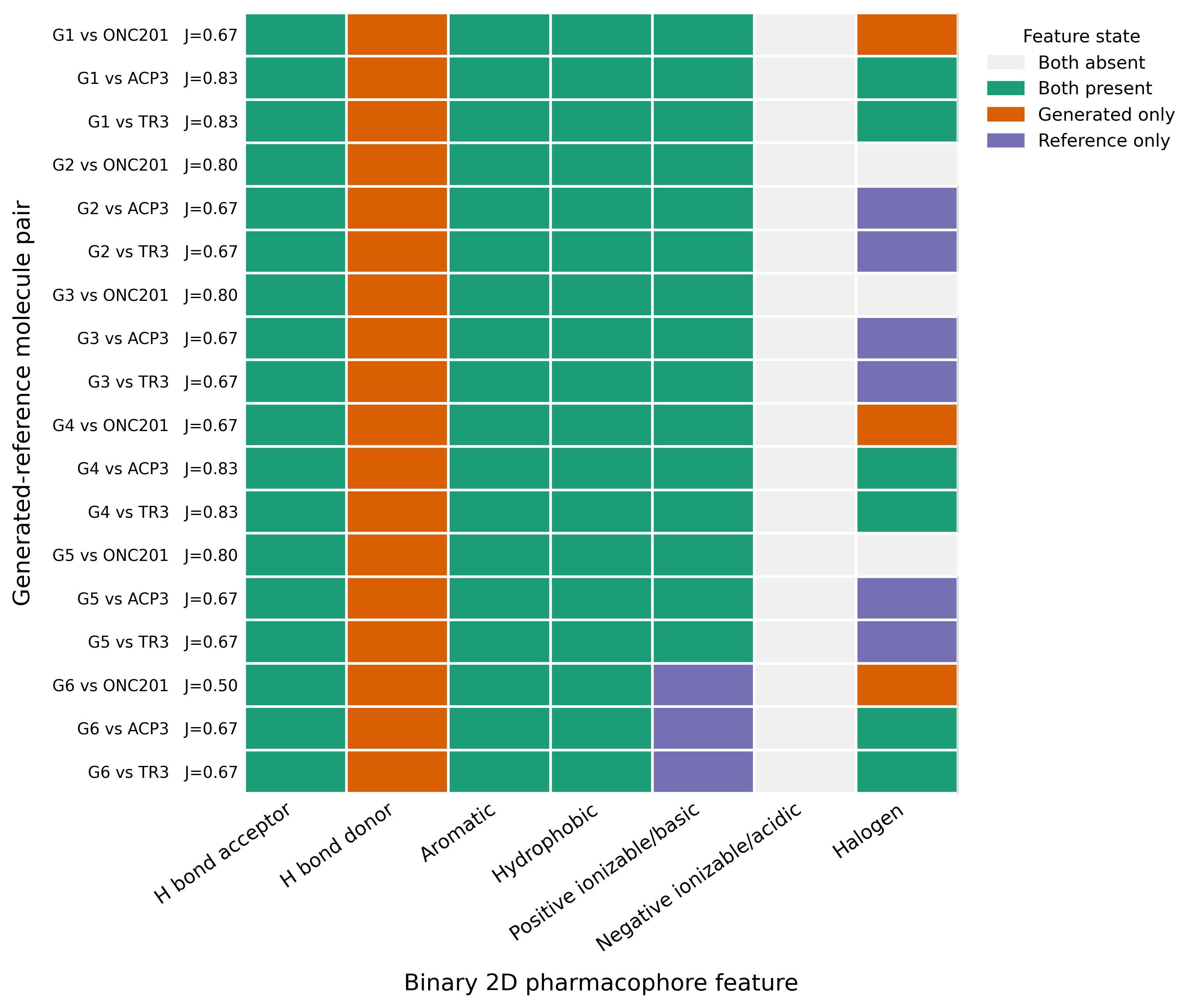}
    \caption{Binary pharmacophore comparison between SpectralMol-generated molecules and reference ClpP activators. Each row corresponds to a generated--reference molecule pair, and each column represents a binary pharmacophore feature. Green indicates that the feature is present in both molecules, orange that the feature is present only in the generated molecule, purple that the feature is present only in the reference molecule, and light gray that the feature is absent in both molecules. The Jaccard coefficient ($J$) reports the pairwise similarity between the generated and reference binary pharmacophore fingerprints. G1 to G6 correspond to generated molecules in Figure \ref{molecules} from left to right and top to bottom.}
    \label{heatmap}
\end{figure}

To quantify the shared relevant motifs, each molecule was encoded using a rule-based binary 2D pharmacophore feature representation. 
The selected pharmacophore features correspond to the general pharmacophoric interaction classes used in established 2D pharmacophore fingerprinting schemes as proposed by Gobbi et al. \cite{gobbi1998genetic}. 
To each feature a boolean value was assigned to encode present or absent using SMARTS-based substructure rules. 
Pairwise similarity between generated and reference molecules was computed using the Jaccard coefficient, defined for two binary feature vectors as the ratio between the number of shared present features and the number of features present in either molecule \cite{jaccard1901distribution,willett1998chemical}.

Figure \ref{heatmap} shows the feature-level comparison in the binary pharmacophore. The generated molecules retain a substantial fraction of the interaction-relevant pharmacophoric profile of the reference ClpP activators. Across all generated--reference pairs, the Jaccard similarity ranges from $J=0.50$ to $J=0.83$, with most comparisons falling between $J=0.67$ and $J=0.83$. The most consistently conserved features are the hydrogen-bond acceptor, aromatic, and hydrophobic features, which are shared across all generated--reference comparisons. The positive ionizable/basic feature is also broadly conserved for G1--G5, whereas G6 lacks this feature relative to all three reference ligands.
The hydrogen-bond donor feature is systematically present in the generated molecules but absent from the reference ligands under the applied SMARTS rules, indicating that the generated candidates introduce an additional donor-type interaction motif rather than simply reproducing the reference pharmacophore pattern. In contrast, the negative ionizable/acidic feature is absent from both generated and reference molecules across all comparisons, indicating that acidic functionality is not a distinguishing feature in this set. Halogen features vary across reference-dependent comparisons. 

Overall, these results indicate that SpectralMol does not merely reproduce the reference molecules, but generates candidates that preserve key ClpP-relevant pharmacophoric elements while introducing controlled variations in substituent chemistry.

\section{Conclusion}

We introduced SpectralMol, a training-free evolutionary framework for multi-objective molecular generation in which the genotype is a compact matrix of Fourier coefficients projected through a fixed basis to produce position-wise latent vectors for SELFIES-constrained decoding. 
Combined with chemically informed token embeddings and an NSGA-II evolutionary algorithm, SpectralMol operates in a compressed parameter space (in the order of thousands Fourier coefficients) without requiring any pre-training stage.
We validated the performance of SpectralMol with a two stage benchmark strategy. First, we compared standardized benchmarks for de-novo molecules generation. SpectralMol achieved overall scores comparable to GraphGA model using GuacaMol optimization tasks. The results showed a task-dependent profile that favors SpectralMol on multi-parameter optimization objectives. 
Second we tested a more realistic ClpP-targeted drug-discovery scenario reproducing Experiment 2 from Guo et al. \cite{guo2024saturn}. Here docking affinity was optimized together with other two drug-relevant features: quantitative drug likeliness (QED) and synthetic accessibility (SA).
SpectralMol generated more unique scaffolds and docking hits as the reinforcement-learning baseline benchmark under a matched oracle-call budget, while preserving comparable QED and molecular weight at the cost of slightly higher SA scores.
Several SpectralMol-generated candidates additionally exhibit architectural motifs consistent with known ClpP activators, suggesting that the diversity gain does not come at the cost of chemical plausibility.

Several limitations remain and define natural directions for further work. The hand-coded vocabulary and embedding construction may miss subtle chemical patterns that learned embeddings capture from large datasets. The frequency-mode ablation supports the mechanistic role of structured Fourier modes, but it also indicates that the balance between low- and high-frequency contributions is task dependent and should be tested further on additional objectives and alternative smooth bases such as wavelets or Gaussian processes over sequence positions. 
More broadly, the results support the view that frequency-controlled evolutionary dynamics provide an interpretable training-free route for multi-objective molecular design, and that this class of representation deserves further investigation as a complement to learned latent-space methods in the de-novo design of small molecules.

\paragraph{Opensource code}
The libraries of the software used is here:  \href{https://github.com/paoloGuida/molevoDrugDiscovery}{GitHub Repository}

\bibliography{references}
\bibliographystyle{unsrt}

\end{document}